\documentclass[10pt,twocolumn,letterpaper]{article}

\usepackage[pagenumbers]{cvpr} 

\usepackage{amsmath, amssymb, amsthm}
\usepackage{mathtools}
\usepackage{bbm}
\usepackage{dsfont}

\newcommand{\bsh}{\boldsymbol{h}}
\newcommand{\bsi}{\boldsymbol{i}}

\newcommand{\bsB}{\boldsymbol{B}}

\newcommand{\bsI}{\boldsymbol{I}}

\newcommand{\bsK}{\boldsymbol{K}}

\newcommand{\bsQ}{\boldsymbol{Q}}

\newcommand{\bsS}{\boldsymbol{S}}

\newcommand{\bsV}{\boldsymbol{V}}
\newcommand{\bsW}{\boldsymbol{W}}

\newcommand{\calB}{{\mathcal{B}}}

\newcommand{\calO}{{\mathcal{O}}}

\newcommand{\bbR}{\mathbb{R}}

\theoremstyle{plain}

\theoremstyle{definition}

\theoremstyle{remark}

\newcommand{\tr}{^\top}

\def\[#1\]{\begin{equation}\begin{aligned}#1\end{aligned}\end{equation}}

\usepackage[acronym,nowarn,section,nogroupskip,nonumberlist]{glossaries}

\glsdisablehyper{}
\newcommand{\acaps}[1]{{\scshape #1}}
\newacronym[\glslongpluralkey={Conditional Neural Processes}]{cnp}{\acaps{cnp}}{Conditional Neural Process}

\newacronym{mil}{\acaps{mil}}{multiple instance learning}
\newacronym{cnn}{\acaps{cnn}}{convolutional neural network}
\newacronym{roi}{\acaps{roi}}{region of interest}
\newacronym{wsi}{\acaps{wsi}}{whole slide image}
\newacronym{acc}{\acaps{acc}}{accuracy}
\newacronym{auc}{\acaps{auc}}{area under the ROC curve}
\newacronym{nll}{\acaps{nll}}{negative log-likelihood}
\newacronym{ece}{\acaps{ece}}{expected calibration error}
\newacronym{pma}{\acaps{pma}}{Pooling by Multi-head Attention}
\newacronym{mc}{\acaps{mc}}{monte carlo}
\newacronym{ssl}{\acaps{ssl}}{self-supervised learning}
\usepackage{cuted}

\usepackage{algorithm}
\usepackage{algorithmic}

\usepackage{fancyvrb}
\usepackage{fvextra}
\usepackage{csquotes} 
\usepackage{adjustbox}
\usepackage{spverbatim}
\usepackage{duckuments}
\usepackage{multicol}

\usepackage[usenames,dvipsnames]{xcolor}

\usepackage{graphicx}

\usepackage{booktabs, array}
\usepackage{multirow}
\usepackage{colortbl}
\definecolor{LightCyan}{rgb}{0.88,0.95,1.0}

\newsavebox\CBox 
\newcommand{\spm}[1]{\scriptstyle{\pm#1}}

\usepackage{subcaption}
\usepackage{duckuments}
\usepackage{booktabs}
\usepackage{arydshln}
\usepackage[dvipsnames]{xcolor}
\usepackage{color, colortbl}

\usepackage{amssymb}
\usepackage{pifont}

\newcommand{\cmark}{\ding{51}}%
\newcommand{\xmark}{\ding{55}}

\definecolor{cvprblue}{rgb}{0.21,0.49,0.74}
\usepackage[pagebackref,breaklinks,colorlinks,citecolor=cvprblue]{hyperref}

\def\paperID{14054} 
\def\confName{CVPR}
\def\confYear{2024}

\title{Slot-Mixup with Subsampling: A Simple Regularization for WSI Classification}

\author{Seongho Keum$^{1}$ \hspace{0.05in} Sanghyun Kim$^{1}$ \hspace{0.05in} Soojeong Lee$^{2}$ \hspace{0.05in} Juho Lee$^{1,3}$ \\
$^{1}$ KAIST, $^{2}$ Sungkyunkwan University, $^{3}$ AITRICS \\
{\tt\small \{shkeum,nannullna,juholee\}@kaist.ac.kr, \{drlisa\}@g.skku.edu}
}

\begin{document}
\maketitle
\begin{abstract}
\Gls{wsi} classification requires repetitive zoom-in and out for pathologists, as only small portions of the slide may be relevant to detecting cancer. Due to the lack of patch-level labels, \gls{mil} is a common practice for training a \gls{wsi} classifier. One of the challenges in \gls{mil} for \glspl{wsi} is the weak supervision coming only from the slide-level labels, often resulting in severe overfitting. In response, researchers have considered adopting patch-level augmentation or applying mixup augmentation, but their applicability remains unverified. Our approach augments the training dataset by sampling a subset of patches in the \gls{wsi} without significantly altering the underlying semantics of the original slides. Additionally, we introduce an efficient model (\emph{Slot-MIL}) that organizes patches into a fixed number of slots, the abstract representation of patches, using an attention mechanism. We empirically demonstrate that the subsampling augmentation helps to make more informative slots by restricting the over-concentration of attention and to improve interpretability. Finally, we illustrate that combining our attention-based aggregation model with subsampling and mixup, which has shown limited compatibility in existing \gls{mil} methods, can enhance both generalization and calibration. Our proposed methods achieve the state-of-the-art performance across various benchmark datasets including class imbalance and distribution shifts.
\end{abstract}

\section{Introduction}
\label{main:sec:introduction} 

\Acrfull{mil}~\citep{dietterich1997solving,maron1997framework} is a variant of weakly-supervised learning where a unit of learning problem is a bag of instances. A bag in \gls{mil} usually contains different numbers of instances, and a label is only assigned to the bag level and unknown for the instances in the bag. Many real-world problems can be formulated as \gls{mil}, including drug-activity prediction~\citep{dietterich1997solving}, document categorization~\citep{andrews2002support}, point-cloud classification~\citep{wu20153d}, and medical image processing~\citep{li2021dual}.

One of the main challenges in \gls{mil} is that we are often given a limited number of labeled bags for training in real-world applications. Consider a \gls{wsi}\footnote{Following words are interchangeable throughout the paper: \gls{wsi} and bag; patch and instance.} classification problem, which is our primary interest in this paper, and a popular example of \gls{mil} in the medical imaging domain. A single \gls{wsi} (bag) usually contains more than tens of thousands of patches (instances), but the labels are only given to the \gls{wsi} level (\emph{i.e.}, whether a \gls{wsi} includes patches corresponding to disease), so the total number of labels given is way smaller than it is needed to train a deep learning model to process tons of patches. Moreover, as for the medical imaging domain, it is common that the \glspl{wsi} from a certain type of disease are scarce in training data, leading to the class imbalance problem. Due to these problems, a na\"ive method for \gls{wsi} classification is likely to suffer from severe overfitting or failure in distribution shifts.

For a usual classification problem, there are several data augmentation techniques to reduce overfitting and thus improve generalization performances. Following the standard protocol in image classification, combining simple augmentations such as rotating, flipping, and cutting on patches of \gls{wsi} might be a valid choice \citep{hendrycks2019augmix}. Mixup augmentation~\citep{zhang2017mixup} might be another option, where a classifier takes a convex combination of two instances as an input and a convex combination of corresponding labels as a label for training. 

Nevertheless, such augmentation techniques are not trivial to apply for \gls{wsi} classification. Augmenting every patch in a \gls{wsi} is highly burdensome due to the massive patch count in a \gls{wsi}. 
Moreover, due to the high cost of processing numerous patches in \glspl{wsi}, it is common to first encode the patches to the set of features using a pre-trained encoder from common image datasets such as ImageNet~\citep{deng2009imagenet}, so the augmentations directly applied to patches are not feasible in such settings. To overcome these innate restrictions, previous works tried to suggest several augmentations specialized for \gls{wsi}. For example, {DTFD-MIL}~\citep{zhang2022dtfd} inflated the number of bags by splitting a \gls{wsi} into multiple chunks and assigning the same label as the original \gls{wsi}. In order to unify the number of patches per \gls{wsi} while not losing information, {RankMix}~\citep{chen2023rankmix} introduced an additional patch classifier which is used as a guidance to choose important patches from \glspl{wsi}, so that they can be selectively used for the mixup augmentation. Although both aforementioned methods tackled the innate problem of \gls{wsi}, they require either knowledge distillation or two-stage training to stabilize training and achieve better performance.

We propose a slot-attention based \gls{mil} method which aggregates patches into a fixed number of slots based on the attention mechanism~\citep{vaswani2017attention,lee2019set,locatello2020object}. As the varying number of patches are summarized into an identical number of slots for every \gls{wsi}, we can directly adopt mixup on slots. Also, we revisit the subsampling augmentation in \gls{wsi} and empirically prove that subsampling helps to mitigate overfitting by involving more crucial patches for decision-making. By integrating subsampling and mixup into our proposal, we can achieve the \emph{state-of-the-art} performance across various datasets. We solved the innate problem of \gls{mil} for \glspl{wsi} through a simple method that does not require any additional process and, thus, is easily applicable.

Our contributions can be outlined as follows:
\begin{itemize}
    \item We propose a slot-based \gls{mil} method, a computationally efficient pooling-based model for \gls{wsi} classification, exhibiting superior performance with fewer parameters.
    \item We unveil the power of subsampling, which surpasses other previous augmentations suggested for \gls{mil}. We also provide a detailed analysis of the effect of subsampling in regularizing attention scores.
    \item As our slot-based method aggregates patches into a fixed number of slots that well represent the \gls{wsi} with the help of subsampling, we can directly adopt mixup to slots. It does not require any extra layer or knowledge distillation, which was essential in previous methods. By doing so, our method obtains SOTA performance with better calibrated predictions.
\end{itemize}

\section{Related Work}
\label{main:sec:related}

\subsection{Deep MIL Models}
To classify a \gls{wsi}, aggregating entire patches into a meaningful representation is crucial. The simplest baseline is to use mean~\citep{pinheiro2015image} or max pooling operations~\citep{feng2017deep} under the assumption that the average or maximum representation across instances can effectively represent the whole bag. ABMIL~\citep{ilse2018attention} first adopted attention mechanisms~\citep{bahdanau2014neural,luong2015effective} in \gls{mil} classification problems and proposed the gated attention to yield instance-wise attention scores with additional non-linearities. However, ABMIL does not consider how patches are related and interact with each other. DSMIL~\citep{li2021dual} utilized two streams of networks for effective classification. The first network finds a critical instance using max pooling, and the second network constructs the bag's representation based on attention scores between the critical instance and the others. TRANSMIL~\citep{shao2021transmil} proposed the TPT module which consists of two transformer layers~\citep{vaswani2017attention} and a positional encoding layer. It adopted self-attention for the first time while trying to reduce the complexity by adopting Nystrom-attention~\citep{xiong2021nystromformer}. ILRA~\citep{xiang2022exploring} used learnable parameters to aggregate patches into smaller subsets and re-expand to patches based on the attention. It repeats this process to acquire meaningful features. 

\subsection{Mixup}
Mixup~\citep{zhang2017mixup} is a simple augmentation strategy where one interpolates pixels of two images and corresponding labels with the same ratio. As mixup provides a smoother decision boundary from class to class, it improves generalization and reduces the memorization of corrupted labels. While simple to implement, mixup has been reported to improve classifiers across various applications, especially for imbalanced datasets or datasets including minority classes~\citep{galdran2021balanced,hwang2022selecmix}. Manifold mixup~\citep{verma2019manifold} adopts the idea of mixup into the feature level and further improves classification performance. As a typical \gls{mil} model utilizes pre-extracted features, mixup in \gls{mil} context means manifold mixup to be precise.

\subsection{Augmentations in WSI Classification}

Recent papers have focused on developing augmentation methods to enhance the limited number of \glspl{wsi} and proposed techniques to prevent overfitting. ReMix~\citep{yang2022remix} uses $k$-means clustering in order to select important patches from a \gls{wsi}. Then, it creates new bags by mixing important patches from different \glspl{wsi} considering Euclidean distance. However, this method can only be applied between \glspl{wsi} with the same label. DTFD-MIL~\citep{zhang2022dtfd} creates pseudo-bags by splitting \gls{wsi} into a subset of patches and assigns the same labels as the original one. When training its first-tier model with pseudo-bags, it selects the crucial patches based on the gradient~\citep{selvaraju2016grad}. Then, with those selected patches, it trains the second-tier model. Adopting manifold mixup directly is not applicable as the size of \glspl{wsi} varies per slide. RankMix~\citep{chen2023rankmix} solves this problem by adding linear layers, so-called \emph{teacher}, on existing models to distinguish important patches. With the help of this layer, we can now unify the number of patches per \glspl{wsi}. It attempts augmentation between \glspl{wsi} with different labels for the first time and proves the effectiveness through performance gain. Nevertheless, it needs pre-training for the teacher model, and its performance without self-training shows little margin.

Orthogonal to mixup, there have been some studies~\citep{combalia2018monte, breen2023efficient} that sample essential patches for classification as utilizing whole patches is computationally heavy. Although they highlighted that random subsampling might improve generalization, they did not clearly explain the correlation between subsampling and attention-based models. 
We revisit this simple but important idea and adapt it to our model.

\begin{table}[t]
\centering
\resizebox{1.0\linewidth}{!}{%
\begin{tabular}{@{}cccc@{}}
\toprule
Model              & Intra & Inter &  Requirements \& Limitations \\ 
\midrule
REMIX~\citep{yang2022remix}             & \color{red}\xmark      & \color{RoyalBlue}\cmark               & Within same label only \\
MIXUP-MIL~\citep{gadermayr2022mixup}          & \color{RoyalBlue}\cmark      & \color{red}\xmark               & Little gain in performance \\
DTFD-MIL~\citep{zhang2022dtfd}           & \color{RoyalBlue}\cmark      & \color{red}\xmark               & Knowledge distillation  \\
RANKMIX~\citep{chen2023rankmix}            & \color{red}\xmark            & \color{RoyalBlue}\cmark         & Pre-training for teacher model  \\ 
\midrule
{\color{blue}Subsampling} + Slot-Mixup (Ours)   & \color{red}\xmark            & \color{RoyalBlue}\cmark         & None of above           \\
\bottomrule
\end{tabular}%
}
\caption{Comparison of augmentation methods for \gls{mil} classification. Intra means augmentation within a single \gls{wsi}, and Inter means augmentation between \glspl{wsi}. Models are labeled considering where the main augmentation occurs.}
\label{main:tab:related_compare}

\end{table}
\section{Method}
\label{main:sec:method}

\subsection{Slot-MIL}
\begin{figure*}[t]
    \begin{subfigure}{0.50\textwidth}
        \centering
        \includegraphics[width=\linewidth]{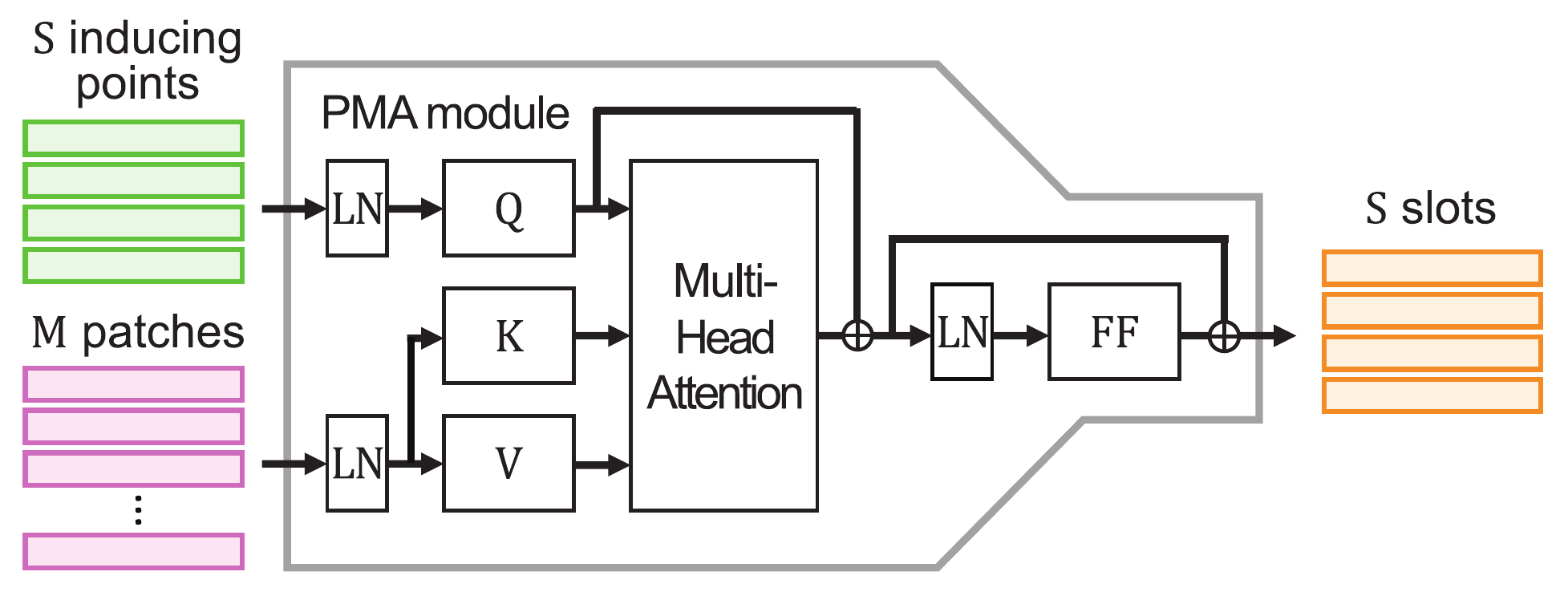}
        \caption{PMA module architecture}
        \label{fig:figure1}
    \end{subfigure}\hfill%
    \begin{subfigure}{0.40\textwidth}
        \centering
        \includegraphics[width=\linewidth]{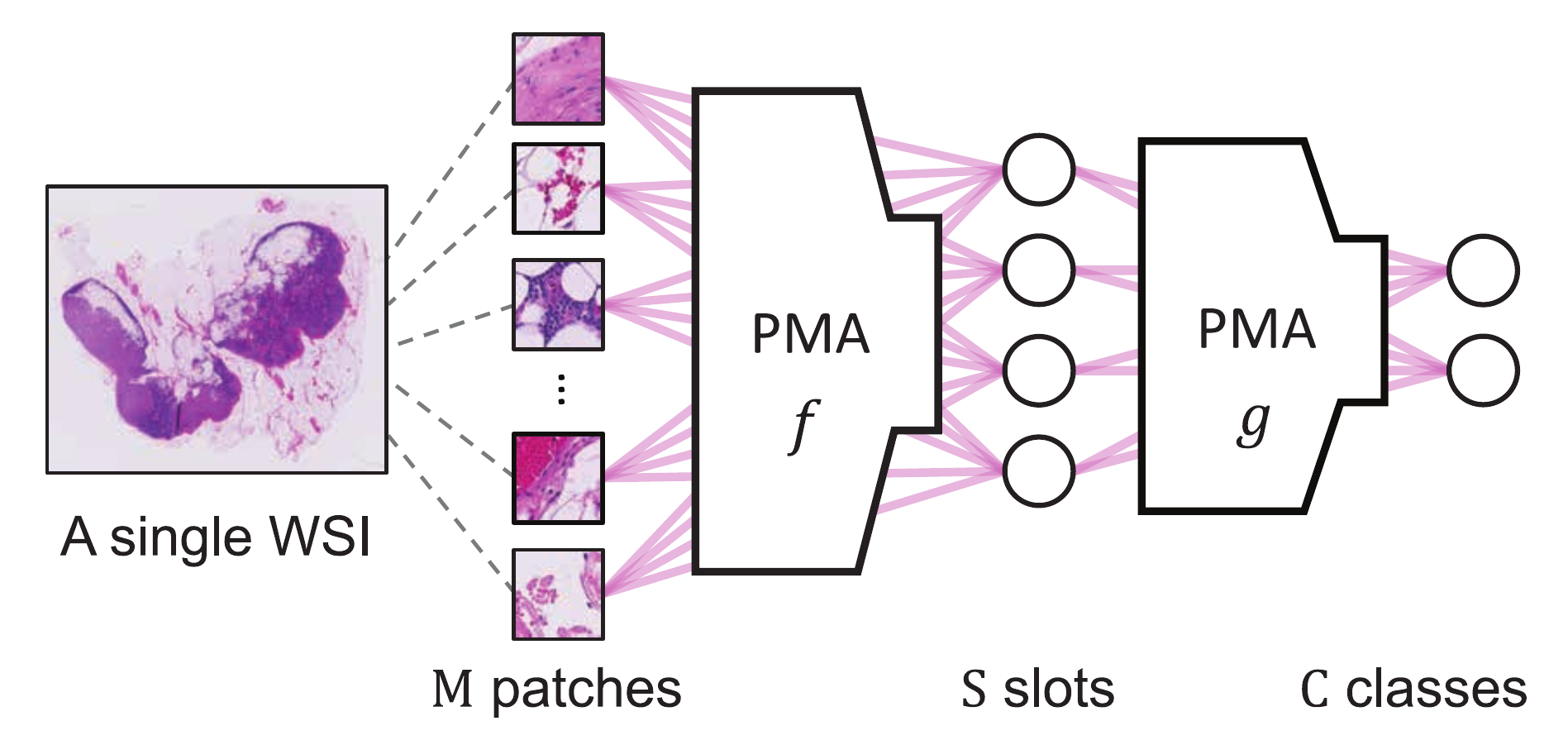}
        \caption{The overall model structure}
        \label{fig:figure2}
    \end{subfigure}
    \caption{(a) The PMA module summarizes $M$ patches into $S$ slots, which is the same number as $S$ learnable vectors called \emph{inducing points}. (b) A single \gls{wsi} is divided into $M$ patches, which then pass through two PMA modules, resulting in logit for each class.}
    \label{fig:both-figures}
\end{figure*}

As it is widely assumed that \gls{wsi} inherently possesses a low-rank structure~\citep{xiang2022exploring}, it is desirable to aggregate a \gls{wsi} into a smaller subset of patches while preserving the underlying semantics. From this perspective, we adopt the idea of inducing points~\citep{lee2019set} and slots~\citep{locatello2020object}, which are the set of learnable vectors used to encode a set of inputs into a fixed-sized feature array. For brevity, we call this module \gls{pma}, following \citet{lee2019set}, with some modifications. Considering the \gls{mil} problems where patches lose absolute position information in the pre-processing stage, we omit positional embedding. 

A bag of $M$ instances is denoted by $\bsB = [\bsh_1, \bsh_2, \dots, \bsh_M]^\top \in \bbR^{M \times d_{\bsh}}$, where $\bsh_i$ is the feature vector of the $i^\text{th}$ instance in the bag computed from a pre-trained encoder and $d_{\bsh}$ is its dimension. We also denote \emph{inducing points} as $\bsI = [\bsi_1, \dots, \bsi_S]^\top \in \bbR^{S \times d_{\bsi}}$, where $S$ is the number of slots with dimension of $d_{\bsi}$. For simplicity, we describe our module with single-head attention, but in practice, we adopt the multi-head attention mechanism as detailed in \citet{vaswani2017attention}. The layer normalization~\citep{ba2016layer, xiong2020layer} is applied to the inducing points and the feature vectors before the attention operation,
\begin{align}
\bar \bsI = \mathrm{LN}(\bsI), \ \bar \bsB = \mathrm{LN}(\bsB),
\end{align}
where $\mathrm{LN}$ is the layer norm operator. Then, we compute the query, key, and value matrices as
\[
\bsQ = \bar \bsI \bsW^Q, \ \bsK = \bar \bsB \bsW^K, \ \bsV = \bar \bsB \bsW^V, 
\]
where $\bsW^Q \! \in \bbR^{d_{\bsi} \times d}$, $\bsW^K \! \in \bbR^{d_{\bsh} \times d}$, and $\bsW^V \! \in \bbR^{d_{\bsh} \times d}$ are learnable parameters, and $d$ is the hyperparameter. Then the updated slot matrix $\bsS$ is computed as
\[
\bsS' &= \mathrm{softmax}(\bsQ\bsK\tr/\sqrt{d})\bsV + \bsQ
\label{eq:softmax-normalization}
\]
\[
\bsS &= \mathrm{MLP}(\mathrm{LN}(\bsS')) + \bsS'.
\label{eq:slot-computation}
\]

As demonstrated in \citet{lee2019set} and \citet{locatello2020object}, the resulting slot matrix $\bsS$ serves as a decent fixed-length summary of the patch features in \glspl{wsi}. Also, a \gls{pma} module is permutation invariant as the slots are not affected by the order of the patches. After summarizing patches into the slots with the first \gls{pma} module $f(\cdot)$, we use another \gls{pma} $g(\cdot)$ to aggregate slots into the classification logits, as illustrated in \cref{fig:figure2}. The only difference from the first \gls{pma} is the dimension of the value matrix, which is set to one, letting the output from $g(\cdot)$ be directly used as classification logits. We name this whole pipeline, using two \glspl{pma} to get logits for a \gls{wsi} classification, \emph{Slot-MIL}. This simple yet powerful model utilizes attention throughout the whole decision process while maintaining relatively fewer parameters and reducing the time complexity from $\calO(M^2)$ to $\calO(SM)$.

A design decision needs to be made when performing the softmax computation in \eqref{eq:softmax-normalization}. In the original softmax operation used in attention mechanisms~\citep{vaswani2017attention,lee2019set}, normalization is applied to the key dimension, meaning that the patch features compete with each other for a slot.  In contrast, in slot attention~\citep{locatello2020object}, the softmax normalization is applied to the query dimension, resulting in slots competing with each other for a patch feature. Interestingly, empirical observations indicate that neither of these two options significantly outperforms the other. As a result, we have explored a hybrid approach, combining both normalization schemes. In practical terms, this involves utilizing key normalization for half of the attention heads and query normalization for the other half. This hybrid strategy has been found to enhance the stability of the training process and, in some datasets, even improve performance.

\subsection{Subsampling}

In this paper, we mainly consider a binary classification problem, where the goal is to build a classifier that takes a bag of feature matrix $\bsB$ as an input and predicts a corresponding binary class $y \in \{0, 1\}$. A common assumption in \gls{mil} is that each instance in a bag has its corresponding latent target variables $y_{1}, \dots, y_{M}$, which are usually unknown. The label of a bag is positive if at least one of its instances is positive, and negative otherwise, i.e., $y = 1 - \prod_{m=1}^M (1 - y_m)$.

\begin{figure*}[t]
    \begin{subfigure}{0.5\textwidth}
        \centering
        \includegraphics[width=\linewidth]{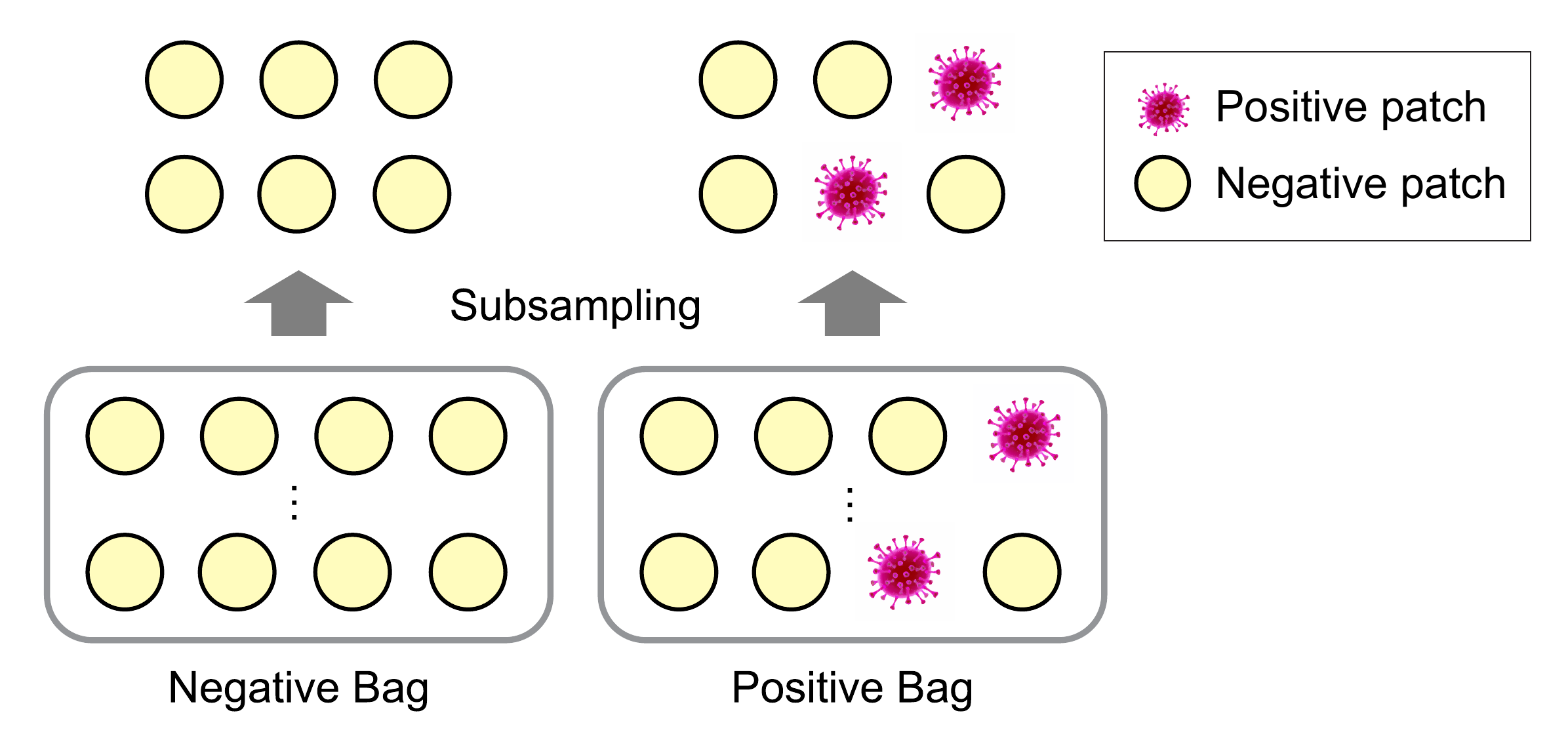}
        \caption{Subsampling in MIL}
        \label{fig:figure3}
    \end{subfigure}\hfill%
    \begin{subfigure}{0.4\textwidth}
        \centering
        \includegraphics[width=\linewidth]{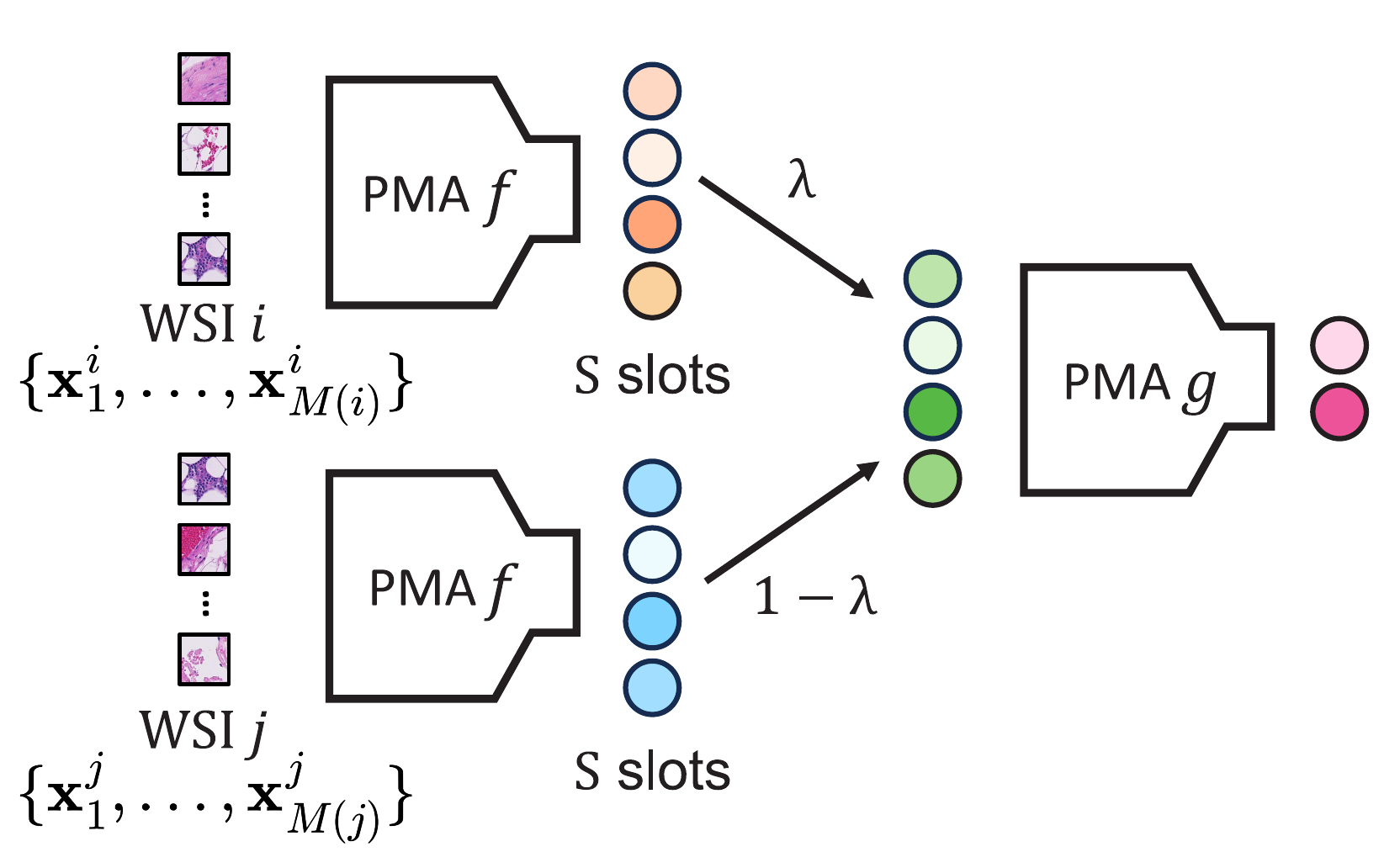}
        \caption{Slot-Mixup}
        \label{fig:figure4}
    \end{subfigure}
    \caption{(a) Subsampling generally maintains the original label when patches are abundant, which is natural in \glspl{wsi}. (b) Illustration of \emph{Slot-Mixup}. As two \glspl{wsi} $i$ and $j$ with different numbers of patches $M(i)$ and $M(j)$, respectively, are summarized into the same number of $S$ slots, applying mixup is straightforward. Two \gls{pma} modules $f$ are identical.}
    \label{main:fig:mixup}
\end{figure*}

One straightforward augmentation method in this scenario is subsampling patches from a bag. \cref{fig:figure3} provides an overview of its procedure, where we first randomly select a subset of patches from a \gls{wsi} without replacement per iteration and then utilize only these chosen subsets for training. It is worth mentioning that several works have touched upon this concept to some extent:
\citet{zhang2022dtfd} proposed a method where a single \gls{wsi} is split into $k$ subsets in order to make pseudo-bags working as augmented data for training a \gls{wsi} classification model;
\citet{chen2023rankmix} also proposed to draw a fixed number of patches from \glspl{wsi}, mainly for the purpose of matching the number of patches between \glspl{wsi} for the mixup augmentation.

We reveal the efficacy of subsampling itself in the context of \gls{wsi} classification or presumably for generic \gls{mil}. A question naturally arises on whether subsampling may alter the semantic of a \gls{mil}; we argue that this is rare, due to the following reasoning. In case of a positive \gls{wsi} slide, for a subsampled bag to be negative, one should subsample all the patches with the negative latent patch labels, which is highly unlikely considering the typical number of patches included in a \gls{wsi}. All the patches are of negative latent labels for a negative \gls{wsi}, so subsampling would not change the slide label anyway. Unlike the previous augmentation strategies, subsampling does not incur additional training costs or extra modifications to the classification model. Furthermore, it further reduces the time complexity of each forward step of training from $\calO(SM)$ to $\calO(pSM)$, where $p \in (0, 1]$ is the subsampling rate.

While we interpret subsampling as a data augmentation technique and thus do not apply it during inference, an alternative perspective is to view subsampling as an instance of dropout~\citep{srivastava2014dropout}. From this standpoint, one might consider performing \gls{mc} inference, where predictions from multiple subsampled inputs are averaged~\citep{gal2016dropout}. However, \gls{mc} inference demands significantly more iterations to match the performance of full-patch inference. Therefore, we don't recommend its use unless better calibration is a specific requirement. Refer to \cref{app:sec:mcinference} for our detailed results on \gls{mc} inference.


\subsection{Slot-Mixup: Mixup Augmentation Using Slots}
As mixup was originally designed for a typical single instance learning problem, applying them for \gls{wsi} classification as well as generic \gls{mil} problems is not straightforward. An apparent problem arises from the fact that, unlike single-instance learning, the unit of prediction here is a bag of patches. There exists an ambiguity of what patches to mix as two slides typically have different numbers of patches. A na\"ive method, as indirectly adopted in \citet{chen2023rankmix}, is to choose a fixed number of patches from two slides. However, as there is no available patch-wise information or annotation, it is difficult to see whether the mixing of arbitrarily chosen patches results in a semantically meaningful augmentation that would enhance the generalization ability of a model trained with them. 

On the other hand, in the case of our Slot-MIL model, given two \glspl{wsi}, the model can summarize them into the identical number of slots encoding the essential information required for classification. Therefore, having the slot summaries of two \glspl{wsi} offers a better option to apply mixup. Specifically, let $\bsS_i$ and $\bsS_j$ be the slots computed from Slot-MIL model in \cref{eq:slot-computation} for $i^\text{th}$ and $j^\text{th}$ slides. Then we draw the mixing ratio $\lambda \sim \calB(\alpha, \alpha)$ and compute the mixed slot set and corresponding label as
\[
&\widetilde{\bsS}_{ij} &= \lambda \bsS_i + (1-\lambda) \bsS_j, \quad \tilde{y}_{ij} = \lambda y_i + (1-\lambda)y_j. 
\]
The application of mixup in this manner offers several advantages in contrast to its prior adaptation for \gls{wsi} classification. By operating within the context of a fixed number of slots, the inherent ambiguity associated with the selection of patches for mixing is effectively eliminated. Concurrently, since these slots serve as concise summaries of the individual instances, the act of mixing slots is anticipated to yield more meaningful augmented data. Additionally, following the principles of manifold mixup, the mixing process occurs at the slot level, eliminating the necessity to recompute patch features each time augmentation is applied during training. This novel approach is termed \emph{Slot-Mixup}, and \cref{fig:figure4} illustrates the procedure. Our experimental results empirically demonstrate its efficacy in addressing the \gls{wsi} classification problem.

Representations of each slot are calculated through learnable inducing points, and thus, their values change during training. In order to mix slots that well represent \glspl{wsi}, we empirically find that starting Slot-Mixup after some epochs is effective. In our experiments, We call this heuristic \emph{late-mix} and study the effect of the number of epochs to start mixup as a hyperparameter $L$.

\subsection{Slot-Mixup with Subsampling}
We combine two augmentation strategies, subsampling and Slot-Mixup, into our Slot-MIL model. As we empirically validate with various benchmarks, two augmentation techniques work well in synergy, resulting in more accurate and better-calibrated predictions. Throughout the paper, we call the combined augmentation strategy of subsampling and Slot-Mixup as SubMix.

\section{Experiments}
\label{main:sec:exp}

\noindent\textbf{Datasets.}
We present experimental findings on three datasets:
\textbf{(1) TCGA-NSCLC}, a subtype classification problem where all slides are of positive cancer. It is a balanced dataset, comprising 528 LUAD and 512 LUSC slides, and ensures slides from the same patient do not overlap between the train and test sets. \textbf{(2) CAMELYON-16}, consisting of 159 positive slides and 238 negative slides, where positive patches occupying less than 10\% of the tissue area in positive slides. Both train and test sets are class-imbalanced. \textbf{(3) CAMELYON-17}, comprising 145 positive slides and 353 negative slides. Distribution shifts exist between train and test splits, as the train and test data are sourced from different medical centers~\citep{litjens20181399}. 
\vspace{0.1in}

\noindent\textbf{Feature extractors.}
The feature extractor based on SimCLR~\citep{chen2020simple} improves the linear separability of patch-level features, posing a challenge in accurately evaluating the contribution of the \gls{mil} methods. Despite potentially inferior performance of features extracted by ImageNet pre-trained ResNet~\citep{he2016deep}, their practicality becomes evident when considering the substantial time required to train SimCLR. Taking these factors into consideration, we conducted experiments using both types of features.
\vspace{0.1in}

\noindent\textbf{Baselines.}
We compare mean/max-pool, ABMIL~\citep{ilse2018attention}, DSMIL~\citep{li2021dual}, TRANSMIL~\citep{shao2021transmil}, and ILRA~\citep{xiang2022exploring} for models without augmentation.
For augmentation methods, we compare our method with DTFD-MIL~\citep{zhang2022dtfd} and RankMix~\citep{chen2023rankmix}. 
\vspace{0.1in}

\noindent\textbf{Evaluation metrics.} 
We made significant efforts to ensure reproducibility and fair comparisons. 
Due to class imbalance and the limited number of slides in benchmarks, we observed that baseline performance is susceptible to the ratio of positive/negative slides in the validation set. Therefore, we opted for a stratified $k$-fold cross-validation. While reporting cross-validation results based on the best validation \gls{auc} has been a convention in the literature, we acknowledge that, given small-sized validation sets, the results can be inconsistent. As a result, we report the average test performance based on the top ten valid \glspl{auc} for each fold. Given that TCGA-NSCLC lacks an official train-test split, we divided the dataset into a train:valid:test ratio of 60:15:25 and performed 4-fold cross-validation. For CAMELYON-16 and CAMELYON-17, which have official train-test splits, we divided the train set into a train:valid ratio of 80:20 for 5-fold cross-validation. 
\vspace{0.1in}

\noindent\textbf{Implementation details.}
We basically follow the hyperparameter setting of \citet{li2021dual} for simplicity. We use Adam optimizer with a learning rate of $10^{-4}$ and trained for 200 epochs. Here, we clarify optimal hyperparameters for SubMix as follows: (number of slots, subsampling rate, late-mix) = ($S$, $p$, $L$) : (16, 0.4, 0.2), (4, 0.2, 0.2), (16, 0.1, 0.2) for CAMELYON-16, CAMELYON-17, and TCGA-NSCLC, respectively. For example, $p=0.1$ indicates the use of 10\% of total patches in a slide per iteration. $L=0.1$ means that we start mixup after 10\% of total epochs.  We find it through grid search, with $p \in \{0.1, 0.2, 0.4\}$, $L \in \{0.1, 0.2, 0.3\}$, and $\alpha \in \{0.2, 0.5, 1.0\}$, after finding the optimal $S$ for Slot-MIL.

\subsection{Subsampling and Mixup}

\subsubsection{Subsampling vs. Other Augmentation Methods}

We first compare subsampling with existing augmentation methods in \cref{main:tab:aug_compare}. We applied DTFD-MIL and RankMix to ABMIL and DSMIL, respectively, following original papers. As they either split a \gls{wsi} or select subset patches by the extra network, it is natural to compare with our subsampling method. With only subsampling, we can achieve performance on par with the above two methods. The result shows the importance of subsampling, which is quite unexplored yet. Optimal results are reported. \Gls{nll} in \gls{mil} area is higher than the natural image area, as they are weakly supervised. The model tends to have high confidence even when it fails to predict correctly. 

\begin{table}[t]
\centering
\resizebox{\linewidth}{!}{%
\begin{tabular}{l|ccc}
\toprule
Dataset & \multicolumn{3}{c}{TCGA-NSCLC} \\
Method & ACC ($\uparrow$) & AUC ($\uparrow$) & NLL ($\downarrow$)  \\ 
\midrule

ABMIL            & {0.832}$\spm{0.039}$ & {0.884}$\spm{0.044}$ & {0.708}$\spm{0.274}$ \\
ABMIL + DTFD-MIL
                 & {0.834}$\spm{0.034}$ & {0.893}$\spm{0.030}$ & {0.980}$\spm{0.178}$ \\
ABMIL + Sub ($p=0.1$)
                 & \cellcolor{LightCyan}\textbf{0.852}$\spm{0.021}$ 
                 & \cellcolor{LightCyan}\textbf{0.920}$\spm{0.021}$ 
                 & \cellcolor{LightCyan}\textbf{0.513}$\spm{0.127}$ \\
ABMIL + Sub ($p=0.2$)
                 & \cellcolor{LightCyan}{0.844}$\spm{0.024}$ 
                 & \cellcolor{LightCyan}{0.914}$\spm{0.023}$ 
                 & \cellcolor{LightCyan}{0.572}$\spm{0.123}$ \\
ABMIL + Sub ($p=0.4$)
                 & \cellcolor{LightCyan}{0.835}$\spm{0.027}$ 
                 & \cellcolor{LightCyan}{0.899}$\spm{0.027}$ 
                 & \cellcolor{LightCyan}{0.640}$\spm{0.103}$ \\

\midrule
DSMIL            & {0.831}$\spm{0.022}$ & {0.897}$\spm{0.021}$ & {0.737}$\spm{0.098}$ \\
DSMIL + RankMix        
                 & {0.820}$\spm{0.026}$ & {0.894}$\spm{0.032}$ & {0.623}$\spm{0.119}$ \\
DSMIL + Sub ($p=0.1$)   
                 & \cellcolor{LightCyan}\textbf{0.853}$\spm{0.026}$ 
                 & \cellcolor{LightCyan}\textbf{0.922}$\spm{0.022}$ 
                 & \cellcolor{LightCyan}\textbf{0.598}$\spm{0.197}$ \\
DSMIL + Sub ($p=0.2$)
                 & \cellcolor{LightCyan}{0.850}$\spm{0.032}$ 
                 & \cellcolor{LightCyan}{0.919}$\spm{0.024}$ 
                 & \cellcolor{LightCyan}{0.623}$\spm{0.173}$ \\
DSMIL + Sub ($p=0.4$)
                 & \cellcolor{LightCyan}{0.845}$\spm{0.025}$ 
                 & \cellcolor{LightCyan}{0.913}$\spm{0.023}$ 
                 & \cellcolor{LightCyan}{0.617}$\spm{0.093}$ \\

\bottomrule
\end{tabular}%
}
\caption{Comparing augmentations in WSI classification. Experiments are conducted based on features extracted by pre-trained ResNet-18.}
\label{main:tab:aug_compare}
\end{table}

\subsubsection{Why Subsampling Works for MIL?}

\begin{figure*}[t]
    \centering
    \includegraphics[width = 0.9\linewidth]{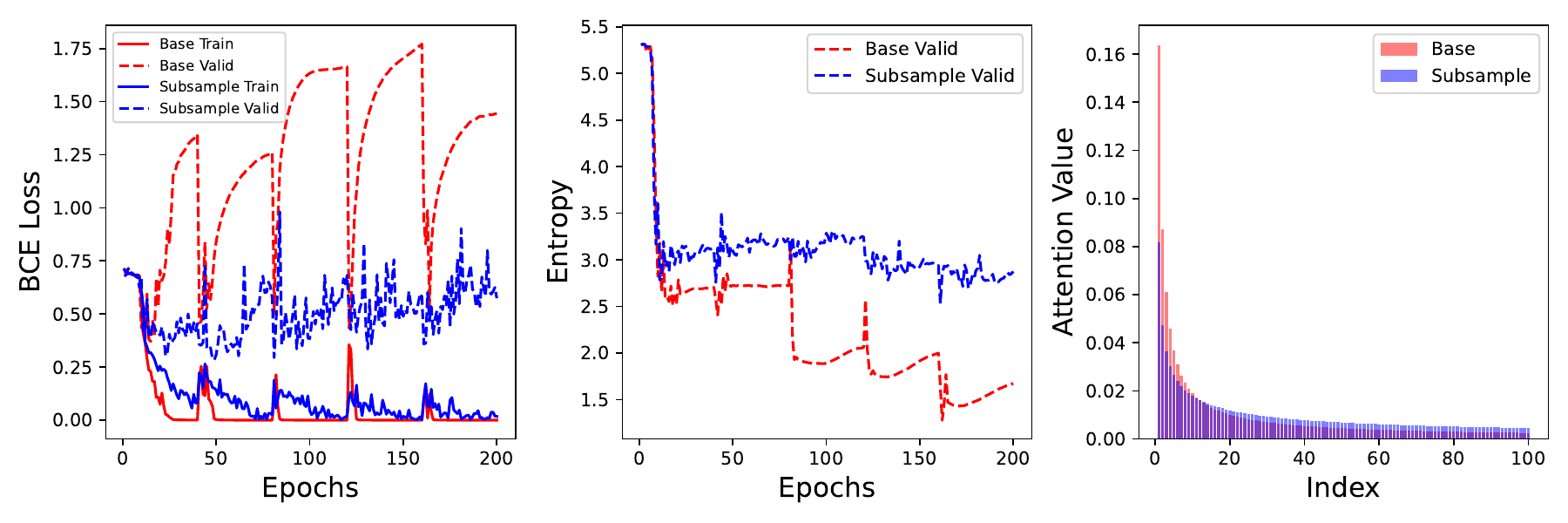}
    \caption{Results of Slot-MIL + Sub on CAMELYON-16. \textbf{(left)} Subsampling improves generalization performance. \textbf{(middle)} Without subsampling, entropy gets smaller as training proceeds, meaning over-concentration of attention to a very limited number of patches. \textbf{(right)} The average of the top 100 attention scores for the test set. Index 1 gets the highest attention score among patches.}
    \label{fig:entropy}
\end{figure*}

Given that a slide-level label can be determined by just one positive patch, it is natural for a model to predominantly concentrate on a very small portion of the slide. However, as the training progresses, attention-based models tend to increasingly focus on smaller subsets, leading to a degradation in generalization. As the combination of positive patches changes through iterations due to subsampling, the model learns to allocate attention to important patches with more equitable weights. As presented in \cref{fig:entropy}, the empirical analysis establishes a correlation between overfitting and attention scores across patches. 

To achieve this, we normalize the sum of attention across all patches to 1 and select the top 100 patches with the highest attention scores. For these patches, we calculate the entropy as $E = -\sum_{a} p(a) \log_2 p(a)$, where $p(a)$ represents the normalized attention scores. A higher entropy signifies a more even distribution of attention scores. Consequently, due to the excessive concentration of attention scores on specific patches, a model that lacks subsampling encounters overfitting during training. Visualizations at the slide-level further strengthen the argument that attention is appropriately distributed to important areas in alignment with pathologists' labels. Moreover, this phenomenon persists irrespective of the types of models and datasets employed.

Through extensive experiments, we observe that adopting nearly any subsampling rate leads to a performance gain compared to not adopting it. This suggests that the advantages of subsampling can be leveraged without requiring an exact prior understanding of the ratio of positive patches. Additional details are in \cref{app:sec:subandreg} and \cref{app:sec:subonbase}.

\subsubsection{Slot-Mixup with Subsampling}

\begin{table}[t]
\centering
\resizebox{\linewidth}{!}{%
\begin{tabular}{l|ccc}
\toprule
Dataset & \multicolumn{3}{c}{TCGA-NSCLC} \\
Method & ACC ($\uparrow$) & AUC ($\uparrow$) & NLL ($\downarrow$)  \\ 
\midrule
Slot-MIL
                 & {0.852}$\spm{0.025}$ & {0.914}$\spm{0.016}$ & {1.001}$\spm{0.202}$ \\
\midrule
+ Sub ($p=0.1$) 
                 & {0.870}$\spm{0.017}$ & {0.929}$\spm{0.019}$ & {0.789}$\spm{0.326}$ \\
+ Sub ($p=0.2$)
                 & \textbf{0.873}$\spm{0.020}$ & \textbf{0.931}$\spm{0.019}$ & {0.751}$\spm{0.333}$ \\
+ Sub ($p=0.4$) 
                 & \textbf{0.873}$\spm{0.020}$ & \textbf{0.931}$\spm{0.023}$ & {0.866}$\spm{0.313}$ \\
\midrule
+ Slot-Mixup ($\alpha=0.2$)
                 & {0.857}$\spm{0.022}$ & {0.915}$\spm{0.019}$ & {0.684}$\spm{0.099}$ \\
+ Slot-Mixup ($\alpha=0.5$)
                 & {0.855}$\spm{0.019}$ & {0.914}$\spm{0.021}$ & {0.686}$\spm{0.161}$ \\
+ Slot-Mixup ($\alpha=1.0$)
                 & {0.856}$\spm{0.020}$ & {0.915}$\spm{0.021}$ & {0.657}$\spm{0.137}$ \\
\midrule
+ SubMix ($p=0.2,\alpha=0.2$)
                 & \cellcolor{LightCyan}{0.871}$\spm{0.019}$ 
                 & \cellcolor{LightCyan}{0.930}$\spm{0.020}$ 
                 & \cellcolor{LightCyan}{0.547}$\spm{0.141}$ \\
+ SubMix ($p=0.2,\alpha=0.5$)
                 & \cellcolor{LightCyan}{0.869}$\spm{0.022}$ 
                 & \cellcolor{LightCyan}\textbf{0.931}$\spm{0.020}$ 
                 & \cellcolor{LightCyan}\textbf{0.496}$\spm{0.135}$ \\
\bottomrule
\end{tabular}%
}
\caption{Comparison of the Subsampling, Slot-Mixup, and SubMix. Experiments are conducted based on features extracted by pre-trained ResNet-18.}
\label{main:tab:submix}
\end{table}

In \cref{main:tab:submix}, we present a comparative analysis of the effects of subsampling, Slot-Mixup, and SubMix when applied to our Slot-MIL model. Subsampling demonstrates an ability to enhance generalization on unseen test data; however, it only marginally improves over-confident predictions compared to the baseline. Conversely, training with mixup exhibits a similar \gls{auc} to the baseline but achieves a lower \gls{nll} due to the generation of intermediate labeled data, contributing to a smoother decision boundary. Combining both techniques, which we refer to as SubMix, allows us to improve generalization performance while maintaining well-calibrated predictions. As a result, for the remaining experiments, we adopt SubMix as our primary augmentation method.

\subsection{WSI Classification}

\subsubsection{WSI Classification Under Distribution Shifts}
\begin{table*}[t]
    \centering
    \small
    \begin{tabular}{l|ccc|cccc}
    \toprule
    Dataset & \multicolumn{7}{c}{CAMELYON-17} \\
    Method & ACC ($\uparrow$) & AUC ($\uparrow$) & NLL ($\downarrow$) 
    & Train & Inference & FLOPs & Model Size \\ 
    \midrule
    Meanpool         & {0.621}$\spm{0.090}$ & {0.661}$\spm{0.034}$ & \textbf{0.622}$\spm{0.110}$
                     & 9.20 & 4.76 & 1,024 & 1,026\\
    Maxpool          & {0.669}$\spm{0.066}$ & {0.602}$\spm{0.033}$ & {0.693}$\spm{0.108}$ 
                     & 9.40 & 4.98 & 1,024 & 1,026\\
    ABMIL            & {0.813}$\spm{0.023}$ & {0.784}$\spm{0.013}$ & {1.056}$\spm{0.488}$ 
                     & 9.59 & 4.79 & 1.32G & 132,483\\
    DSMIL            & {0.735}$\spm{0.044}$ & {0.745}$\spm{0.027}$ & {1.894}$\spm{0.847}$ 
                     & 10.15 & 4.84 & 3.30G & 331,396\\
    TRANS-MIL        & {0.727}$\spm{0.063}$ & {0.739}$\spm{0.059}$ & {2.521}$\spm{0.472}$ 
                     & 36.92 & 9.02 & 50.31G & 2,147,346\\
    ILRA             & {0.716}$\spm{0.027}$ & {0.713}$\spm{0.036}$ & {1.196}$\spm{0.198}$
                     & 27.53 & 7.98 & 9.92G & 1,555,330\\
    Slot-MIL 
                     & \cellcolor{LightCyan}\textbf{0.832}$\spm{0.006}$ 
                     & \cellcolor{LightCyan}{0.813}$\spm{0.016}$ 
                     & \cellcolor{LightCyan}{1.181}$\spm{0.425}$
                     & \cellcolor{LightCyan}12.53 & \cellcolor{LightCyan}5.63 
                     & \cellcolor{LightCyan}5.45G & \cellcolor{LightCyan}1,590,785\\
    \midrule
    \midrule
    DTFD-MIL
                     & {0.793}$\spm{0.026}$ & {0.819}$\spm{0.014}$ & {1.206}$\spm{0.387}$ 
                     & 15.09 & 6.90 & - & -\\
    Slot-MIL + RankMix
                     & {0.823}$\spm{0.032}$ & {0.817}$\spm{0.025}$ & {0.640}$\spm{0.124}$ 
                     & 30.91 & 5.64 & - & -\\
    Slot-MIL + SubMix
                     & \cellcolor{LightCyan}{0.805}$\spm{0.016}$ 
                     & \cellcolor{LightCyan}\textbf{0.835}$\spm{0.015}$ 
                     & \cellcolor{LightCyan}{0.633}$\spm{0.035}$
                     & \cellcolor{LightCyan}22.74 & \cellcolor{LightCyan}5.66 
                     & \cellcolor{LightCyan}- & \cellcolor{LightCyan}-\\
    
    \bottomrule
    \end{tabular}
    \caption{Results on CAMELYON-17 (distribution shifts); Experiments are conducted on the features extracted from the pre-trained ResNet-18. Comparison without augmentation is above the double line, and with augmentation is below the double line. DTFD-MIL is applied to ABMIL following the original paper. Training time and inference time are reported in seconds per epoch. We measure FLOPs with a bag size of 10,000.}
    \label{main:tab:cam17}
\end{table*}

CAMELYON-17 experiences distribution shifts due to the use of different scanners in separate medical centers for training and testing, compounded by variations in data collection and processing methods. Traditional pre-training methods do not effectively address this issue~\citep{wiles2021fine}, and fine-tuning specific layers shows some promising directions~\citep{lee2022surgical}. As subsampling generates diverse in-distribution samples, and mixup leverages distributions from two distinct samples, SubMix shows robust performance on distribution shifts.

Our Slot-MIL model demonstrates the \emph{state-of-the-art} (SOTA) performance, particularly in a distribution-shifted domain, as illustrated in \cref{main:tab:cam17}.  Furthermore, the SubMix augmentation technique significantly enhances performance, affirming the efficacy of our approach even in the absence of pre-training or fine-tuning. Please note that \gls{auc} is recognized as a more robust metric than \gls{acc} as it is not influenced by threshold variations. 

Additionally, when compared to fully attention-based methods like TRANS-MIL and ILRA, Slot-MIL demonstrates 2.2 times faster training time and 1.4 times faster inference time.
Also, the FLOPs of Slot-MIL are significantly smaller, attributed to the absence of positional encoding networks and iterative attention layers which are essential for others. It's noteworthy that our method achieves fast inference time while demonstrating superior performance, highlighting advantages in real-world scenarios.

\subsubsection{Standard WSI Classification}

\cref{main:tab:main} shows the results for standard \gls{wsi} classification, where there are no distribution shifts, as \glspl{wsi} from different scanners exist in both the train and the test sets. Slot-MIL demonstrates SOTA performance in both datasets, regardless of the feature extraction type, when augmentation is not adopted. Additionally, SubMix outperforms RankMix, providing evidence for the effectiveness of attention-based aggregation methods that do not require patch selection for Mixup. Considering that RankMix needs additional training of a teacher model which requires around 2x training complexity than SubMix, our method is efficient and powerful. The results without self-training on RankMix are in \cref{appendix:sec:woteacher} for a fair comparison in terms of complexity. Without self-training, SubMix outperforms RankMix.

\begin{table*}[t]
\centering
\resizebox{\linewidth}{!}{%
\begin{tabular}{l|cc|cc|cc|cc}
\toprule
Dataset & \multicolumn{4}{c|}{CAMELYON-16} & \multicolumn{4}{c}{TCGA-NSCLC}\\
Feature & \multicolumn{2}{c|}{ResNet-50} & \multicolumn{2}{c|}{SimCLR} & \multicolumn{2}{c|}{ResNet-18} & \multicolumn{2}{c}{SimCLR}\\
Method & AUC ($\uparrow$) & NLL ($\downarrow$) & AUC ($\uparrow$) & NLL ($\downarrow$)
& AUC ($\uparrow$) & NLL ($\downarrow$) & AUC ($\uparrow$) & NLL ($\downarrow$)\\
\midrule
Meanpool         & {0.522}$\spm{0.036}$ & {0.971}$\spm{0.064}$ & {0.604}$\spm{0.003}$ & {0.674}$\spm{0.023}$
                 & {0.798}$\spm{0.025}$ & {0.571}$\spm{0.047}$ & {0.972}$\spm{0.010}$ & \textbf{0.232}$\spm{0.055}$\\ 
Maxpool          & {0.783}$\spm{0.022}$ & {0.942}$\spm{0.258}$ & {0.967}$\spm{0.002}$ & {0.353}$\spm{0.147}$
                 & {0.802}$\spm{0.011}$ & {0.572}$\spm{0.023}$ & {0.961}$\spm{0.013}$ & {0.608}$\spm{0.053}$\\
ABMIL            & {0.808}$\spm{0.034}$ & {1.185}$\spm{0.395}$ & {0.972}$\spm{0.002}$ & {0.234}$\spm{0.033}$
                 & {0.884}$\spm{0.044}$ & {0.708}$\spm{0.274}$ & \textbf{0.981}$\spm{0.010}$ & {0.263}$\spm{0.101}$\\
DSMIL            & {0.833}$\spm{0.063}$ & {1.620}$\spm{1.145}$ & {0.968}$\spm{0.008}$ & {0.456}$\spm{0.182}$
                 & {0.897}$\spm{0.021}$ & {0.737}$\spm{0.098}$ & \textbf{0.981}$\spm{0.010}$ & {0.324}$\spm{0.133}$\\
TRANSMIL         & {0.834}$\spm{0.036}$ & {1.654}$\spm{0.326}$ & {0.939}$\spm{0.010}$ & {0.988}$\spm{0.188}$
                 & {0.893}$\spm{0.021}$ & {1.791}$\spm{0.559}$ & {0.974}$\spm{0.009}$ & {0.381}$\spm{0.127}$\\
ILRA             & {0.842}$\spm{0.051}$ & {1.157}$\spm{0.396}$ & {0.973}$\spm{0.007}$ & {0.333}$\spm{0.043}$
                 & {0.901}$\spm{0.028}$ & {0.824}$\spm{0.117}$ & \textbf{0.981}$\spm{0.011}$ & {0.277}$\spm{0.103}$\\

Slot-MIL        & \cellcolor{LightCyan}{0.893}$\spm{0.023}$ 
                & \cellcolor{LightCyan}{1.242}$\spm{0.979}$
                & \cellcolor{LightCyan}{0.972}$\spm{0.007}$ 
                & \cellcolor{LightCyan}{0.294}$\spm{0.065}$ 
                & \cellcolor{LightCyan}{0.914}$\spm{0.016}$ 
                & \cellcolor{LightCyan}{1.001}$\spm{0.202}$ 
                & \cellcolor{LightCyan}\textbf{0.981}$\spm{0.011}$ 
                & \cellcolor{LightCyan}{0.276}$\spm{0.131}$\\
\midrule
\midrule
DTFD-MIL
                 & {0.844}$\spm{0.052}$ & {1.014}$\spm{0.255}$ & \textbf{0.975}$\spm{0.004}$ & {0.292}$\spm{0.035}$
                 & {0.893}$\spm{0.030}$ & {0.980}$\spm{0.178}$ & \textbf{0.981}$\spm{0.011}$ & {0.309}$\spm{0.135}$\\
RankMix
                 & {0.914}$\spm{0.025}$ & {0.525}$\spm{0.087}$ & {0.965}$\spm{0.012}$ & {0.342}$\spm{0.064}$
                 & \textbf{0.932}$\spm{0.021}$ & {0.532}$\spm{0.173}$ & {0.980}$\spm{0.011}$ & {0.283}$\spm{0.014}$ \\
SubMix 
                 & \cellcolor{LightCyan}\textbf{0.921}$\spm{0.020}$ 
                 & \cellcolor{LightCyan}\textbf{0.448}$\spm{0.103}$
                 & \cellcolor{LightCyan}\textbf{0.975}$\spm{0.008}$ 
                 & \cellcolor{LightCyan}\textbf{0.229}$\spm{0.071}$
                 & \cellcolor{LightCyan}{0.931}$\spm{0.020}$ 
                 & \cellcolor{LightCyan}\textbf{0.496}$\spm{0.135}$
                 & \cellcolor{LightCyan}\textbf{0.981}$\spm{0.012}$ 
                 & \cellcolor{LightCyan}{0.248}$\spm{0.105}$\\

\bottomrule
\end{tabular}%
}
\caption{Results on CAMELYON-16 and TCGA-NSCLC. RankMix and SubMix is applied to Slot-MIL, but omitted for brevity. We utilize the open-source features provided by ~\citet{zhang2022dtfd} and ~\citet{li2021dual} for ResNet-50 and SimCLR experiments, respectively.}
\label{main:tab:main}
\end{table*}

\subsection{Further Analysis and Ablation Studies}
We present an in-depth analysis of the hyperparameters governing our Slot-MIL and SubMix methodologies. Further results are detailed in \cref{appendix:sec:latemix}.

\noindent\textbf{Number of slots, $S$.} We empirically demonstrate that a small number of slots is sufficient to capture the underlying semantics of \glspl{wsi}. There is minimal difference in performance when the number of slots exceeds a certain threshold. Consequently, we determine the optimal number of slots as (16, 4, 16) for CAMELYON-16, CAMELYON-17, and TCGA-NSCLC, taking computational efficiency into consideration.
\vspace{0.1in}

\noindent\textbf{Mixup hyperparameter, $\alpha$.} The mixup ratio, $\lambda$, varies during each iteration and is sampled from the beta distribution $\mathcal B(\alpha, \alpha)$. Across different datasets, we observe that values of $\alpha$ greater than 1 degrades performance, as they lead to a more pronounced divergence between the mixed feature distribution and the original distribution.
\vspace{0.1in}

\noindent\textbf{Subsampling rate, $p$.} The subsampling rate is directly related to the proportion of positive patches within positive slides. In the case of TCGA-NSCLC, where positive patches make up approximately 80\% of the dataset, even a small value of $p$ suffices to capture the underlying semantics of the original labels. Conversely, in CAMELYON-16, where positive patches constitute only 10\% of positive slides, larger values of $p$ prove to be effective. 
\vspace{0.1in}

\noindent\textbf{Late-mix parameter, $L$.} Allowing the slots to learn the underlying representations of \glspl{wsi} from the un-mixed original training set is crucial. Hence, the application of mixup after a certain number of epochs becomes particularly essential, especially in the case of CAMELYON-16 and CAMELYON-17. 

\begin{table}[t]
    \label{main:tab:hyper}
    \centering
    \small
    \begin{tabular}{ccc}
        \toprule
        Hyperparam & AUC ($\uparrow$) & NLL ($\downarrow$) \\
        \midrule
        $S=4$  & {0.874}$\spm{0.026}$ & {1.000}$\spm{0.444}$  \\
        $S=8$  & {0.892}$\spm{0.024}$ & {1.221}$\spm{0.445}$  \\
        $S=16$ & {0.893}$\spm{0.023}$ & {1.242}$\spm{0.979}$  \\
        $S=32$ & {0.891}$\spm{0.031}$ & {1.438}$\spm{0.536}$  \\
        \midrule
        $\alpha=0.2$  & {0.919}$\spm{0.021}$ & {0.496}$\spm{0.122}$  \\
        $\alpha=0.5$  & {0.921}$\spm{0.020}$ & {0.448}$\spm{0.103}$  \\
        $\alpha=1.0$  & {0.906}$\spm{0.018}$ & {0.480}$\spm{0.054}$  \\
        \midrule
        $L=0$  & {0.908}$\spm{0.016}$ & {0.572}$\spm{0.074}$  \\
        $L=0.1$  & {0.907}$\spm{0.017}$ & {0.509}$\spm{0.072}$  \\
        $L=0.2$  & {0.921}$\spm{0.020}$ & {0.448}$\spm{0.103}$  \\
        \bottomrule
    \end{tabular}
    \caption{Ablations on CAMELYON-16; We use features extracted by pre-trained ResNet-50. $S$ ablation is performed for Slot-MIL. For the $\alpha$ and $L$ ablation with SubMix, $S$ and $p$ are fixed at 16 and 0.4. When $L$ is set to 0, it indicates that Late-mix is not applied.}
\end{table}

\section{Conclusion}
\label{main:sec:conclusion}

\gls{wsi} classification suffers extreme overfitting due to a lack of data and weak signal coming only from the slide-level label. In order to solve this problem, previous studies tried to suggest augmentations but their approach is either ineffective or complex. Based on our efficient model Slot-MIL, which aggregates patches into informative slots, we can easily apply Slot-Mixup. Also, we uncover the effect of subsampling on the attention-based model in \gls{mil}, which is quite unexplored yet. Utilizing subsampling, and Slot-Mixup concurrently, we achieve SOTA performance in various datasets with calibrated prediction superior to other methods. While it may not be calibrated as well as a natural image, we hope that our model can contribute to diagnosing cancer in real-world applications. As we can unify the number of patches in \glspl{wsi} with subsampling, future research includes mini-batch training which is not investigated well in \gls{mil} for \glspl{wsi}.

\clearpage
\newpage
{
\small
\bibliographystyle{ieeenat_fullname}
\bibliography{main}
}
\clearpage
\newpage
\appendix
\setcounter{page}{1}
\maketitlesupplementary

\section{Subsampling}
\label{app:sec:subandreg}
\subsection{Subsampling and Attention Regularization}
Subsampling plays a crucial role in preventing overfitting by giving more balanced attention weights to important patches. This phenomenon holds true across different models and datasets, making it a general strategy. In TCGA-NSCLC, where positive slides contain more positive patches than CAMELYON-16, we observe higher entropy when subsampling is applied.

\begin{figure}[hbt!]
    \centering
    \includegraphics[width = 0.75\linewidth]{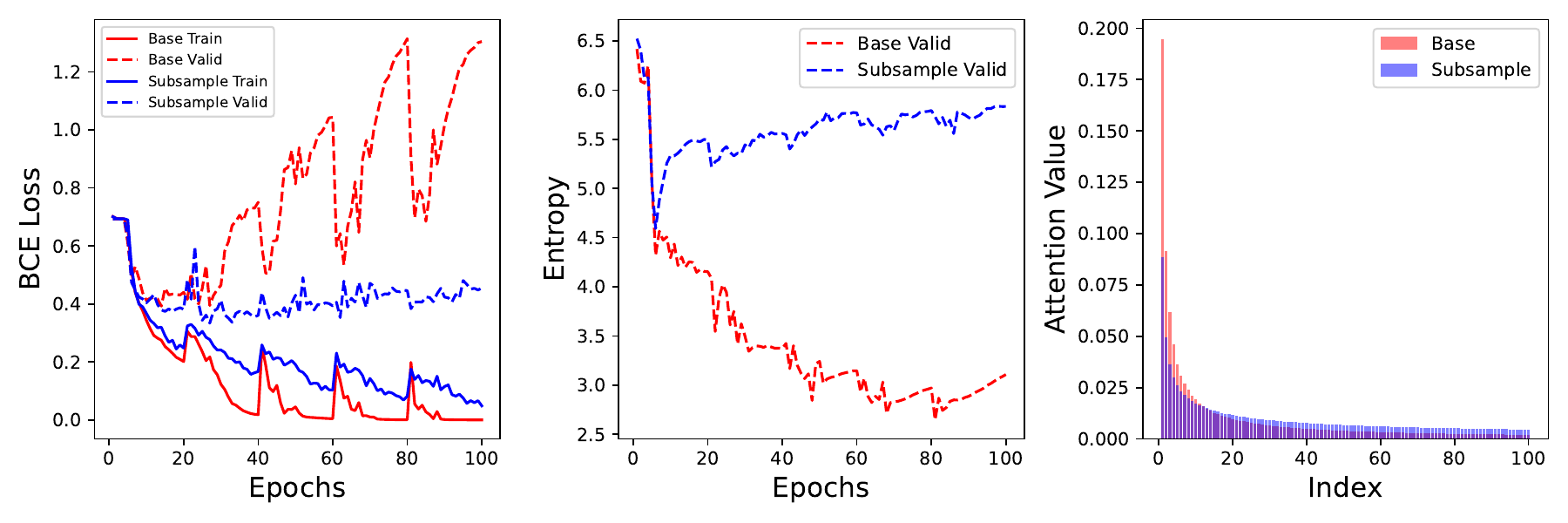}
    \caption{Results of Slot-MIL on TCGA-NSCLC. Baseline doesn't utilize subsampling.}
\end{figure}

\begin{figure}[hbt!]
    \begin{subfigure}{0.48\textwidth}
        \centering
        \includegraphics[width = 0.9\linewidth]{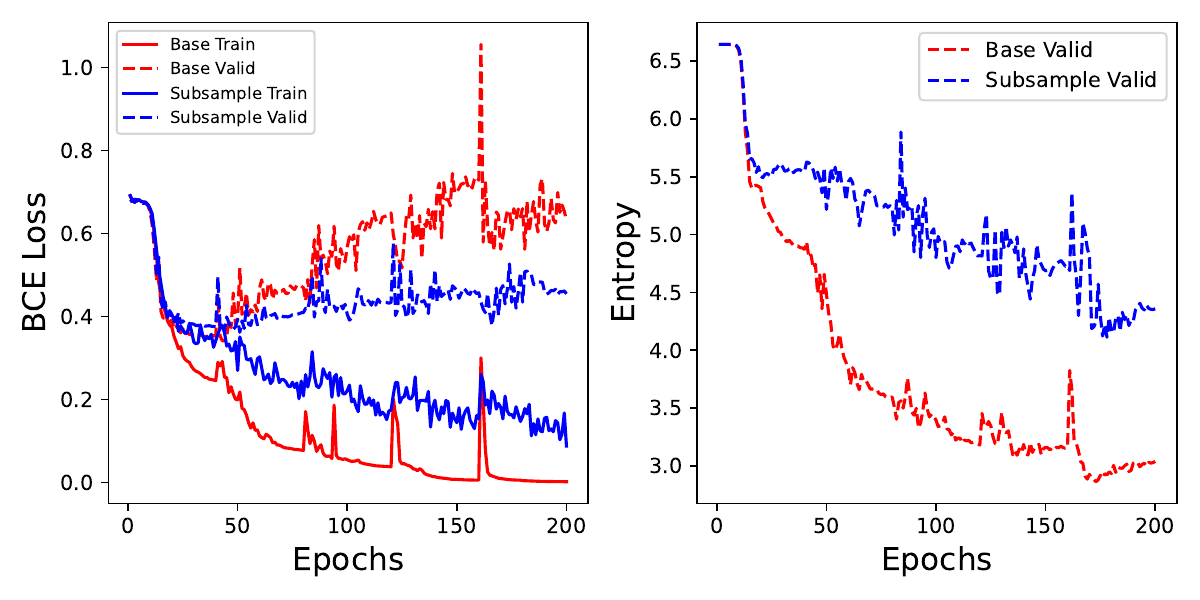}
        \caption{Results of ABMIL on CAMELYON-16.}
    \end{subfigure}
    \begin{subfigure}{0.48\textwidth}
        \centering
        \includegraphics[width = 0.9\linewidth]{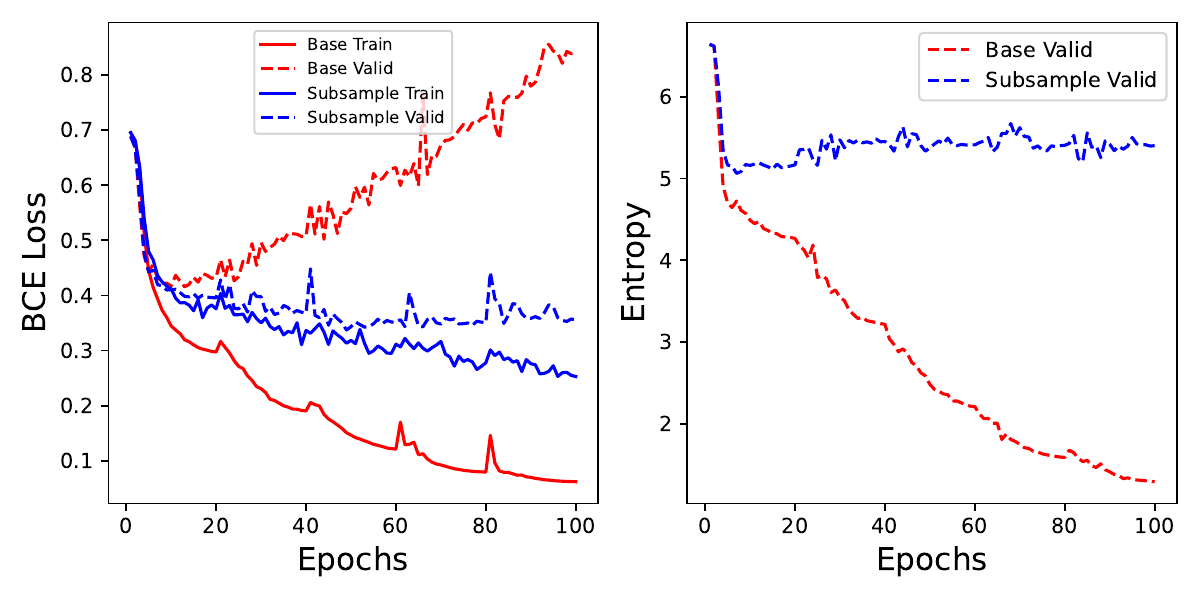}
        \caption{Results of ABMIL on TCGA-NSCLC.}
    \end{subfigure}
    \caption{Subsampling helps to mitigate overfitting in ABMIL.}
\end{figure}

\begin{figure}[hbt!]
    \begin{subfigure}{0.48\textwidth}
        \centering
        \includegraphics[width = 0.9\linewidth]{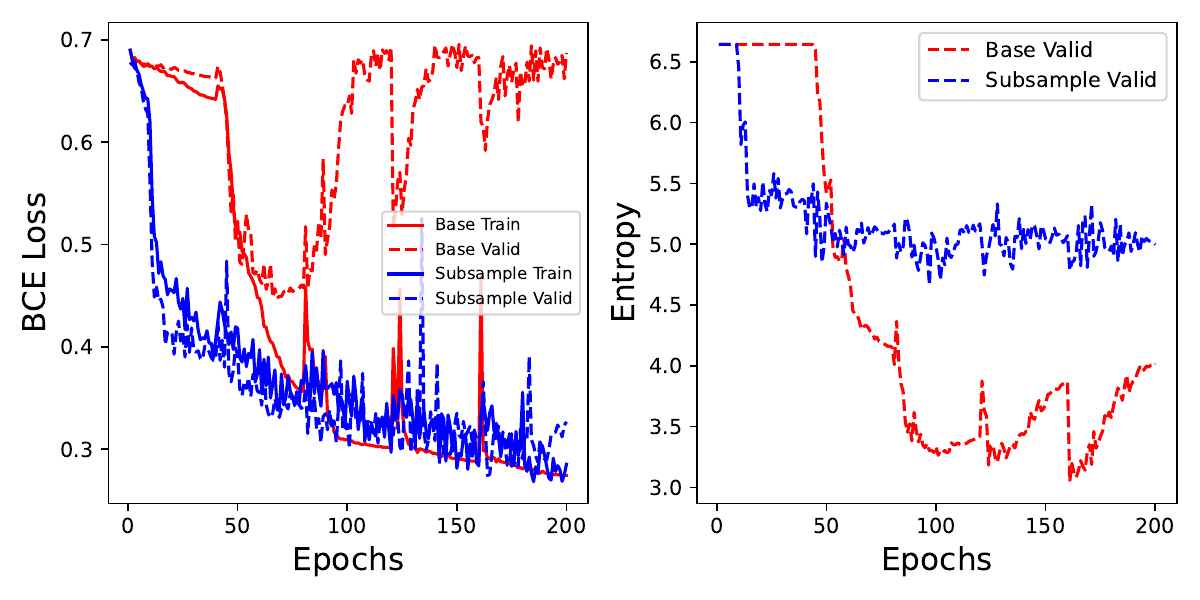}
        \caption{Results of DSMIL on CAMELYON-16.}
    \end{subfigure}
    \begin{subfigure}{0.48\textwidth}
        \centering
        \includegraphics[width = 0.9\linewidth]{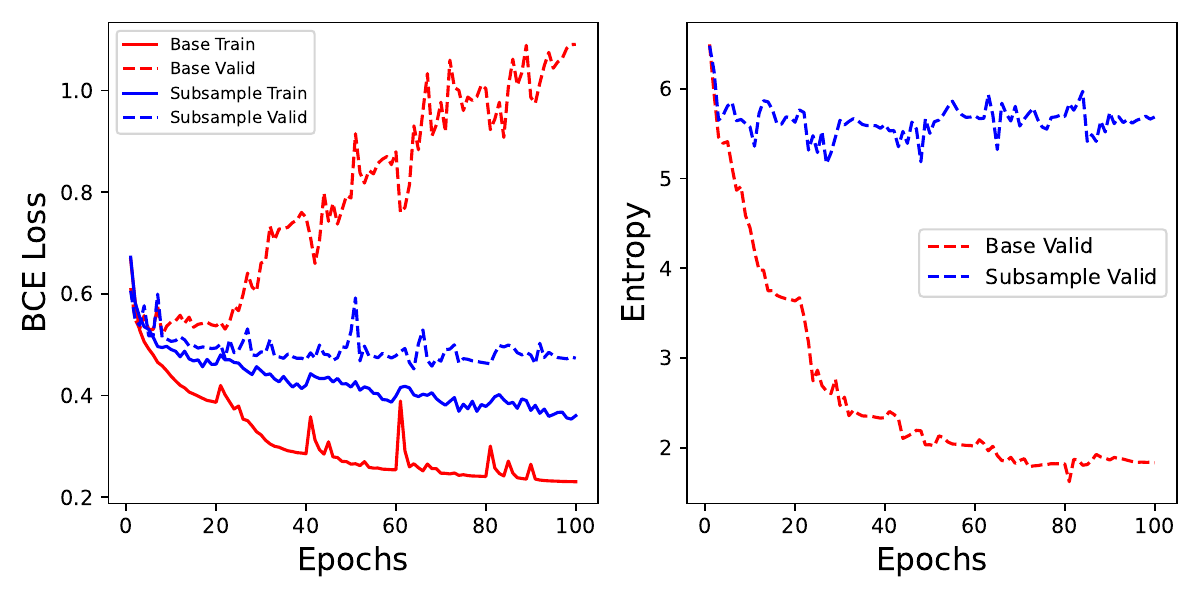}
        \caption{Results of DSMIL on TCGA-NSCLC.}
    \end{subfigure}
    \caption{Subsampling helps to mitigate overfitting in DSMIL.}
\end{figure}

\subsection{Subsampling and Patch-level Annotation}
\label{appendix:sec:visualization}

\begin{figure}[hbt!]
    \centering
    \includegraphics[width=0.8\linewidth]{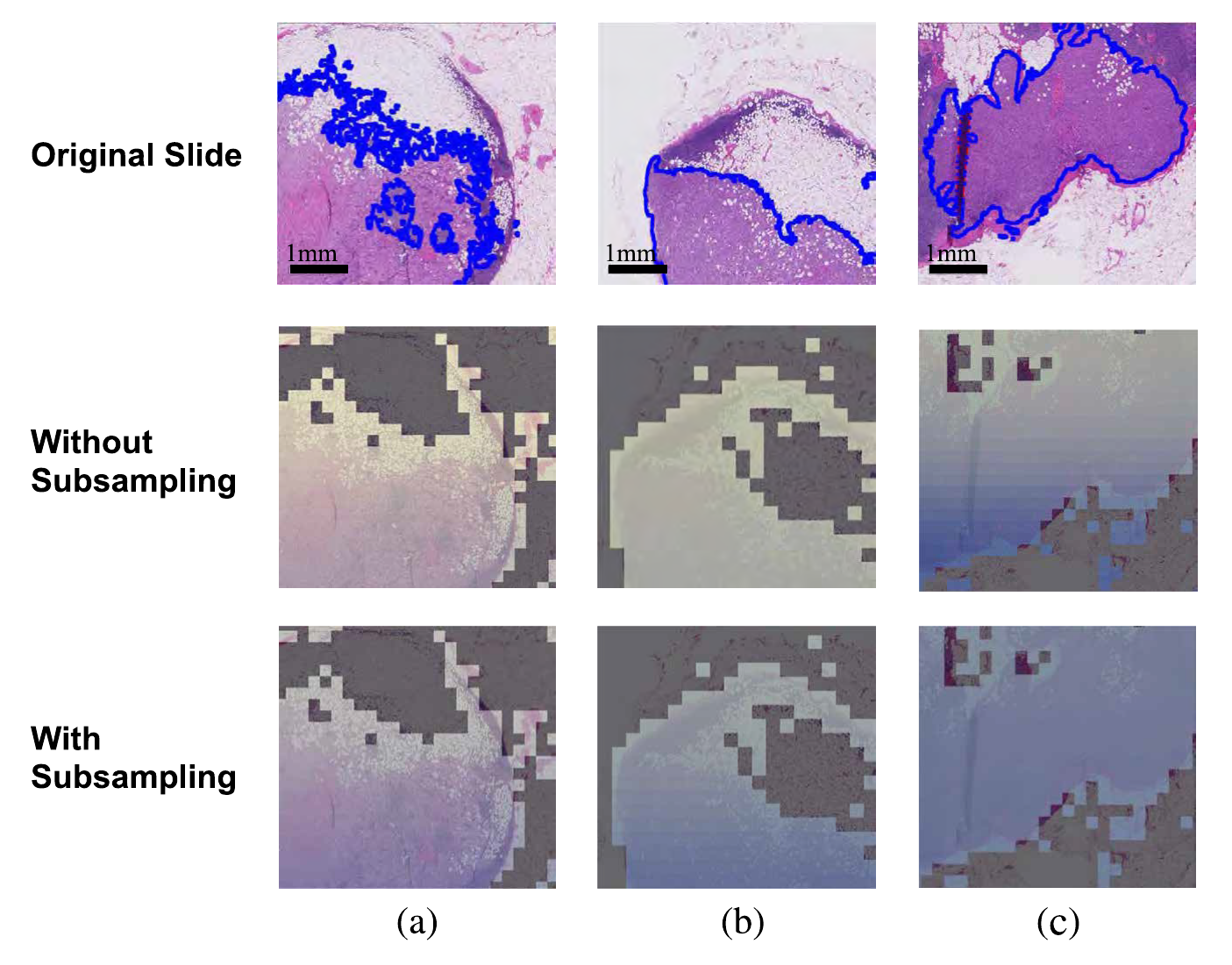}
    \caption{With subsampling, the attention is distributed more evenly, aligning with experts' annotations. \textbf{(a), (b)} With the help of subsampling, Slot-MIL detects tumor area more accurately. \textbf{(c)} Without subsampling, the attention score tends to concentrate on a specific area.}
    \label{fig:attention_tumor}
\end{figure}

\cref{fig:attention_tumor} visually illustrates how slot's attention score is directed towards patches, utilizing the CAMELYON-16 dataset. We choose this dataset as it has patch-level annotation. Experts' annotation is shown with blue lines. Comparing the second row (without subsampling) and the third row (with subsampling) for three representative examples, based on our Slot-MIL, reveals a notable enhancement in tumor area detection upon the application of subsampling. This improvement contributes to the generation of more informative slots. Deeper shades of blue indicate higher attention score, while shades closer to white indicate lower attention score. Gray patches represent the background, which is excluded in the pre-processing stage.

\section{Multi-class classification}
We conduct experiments on a challenging multi-class classification task using the CAMELYON-17 dataset. This dataset encompasses four classes: negative, itc, micro, and macro, categorized based on the size of tumor cells. In our primary analysis, we treated negative and itc as the negative class, while micro and macro were considered positive, facilitating binary classification. This dataset is characterized by class imbalance and distribution shifts, as it involves data collected from multiple hospitals, adding complexity to the problem. In this challenging setting, our Slot-MIL demonstrates relatively decent performance, and the SubMix method significantly enhances accuracy while providing more calibrated predictions, in terms of both \gls{ece} and \gls{nll}. Also, SubMix performs better than just applying subsampling to Slot-MIL, which further strengths the validity of Slot-Mixup which is quiet limited to calibration improvement in binary classification. Additionally, we observe a significantly high Pearson's correlation of $r=0.9313$ ($\text{p-value} = 0.0003$) between \gls{ece} and \gls{nll}. 

\begin{table}[hbt!]
\centering
\begin{tabular}{l|cccc}
\toprule
Method/Dataset  &\multicolumn{4}{c}{CAMELYON-17} \\
& ACC ($\uparrow$) & AUC ($\uparrow$) & NLL ($\downarrow$) & ECE ($\downarrow$)\\ 
\midrule
Meanpool 
& {0.642}$\spm{0.029}$ & {0.601}$\spm{0.024}$ & {0.969}$\spm{0.036}$ & {0.095}$\spm{0.028}$\\
Maxpool
& {0.701}$\spm{0.012}$ & {0.620}$\spm{0.015}$ & {0.894}$\spm{0.010}$ & {0.077}$\spm{0.027}$\\
ABMIL    
& {0.702}$\spm{0.025}$ & {0.677}$\spm{0.024}$ & {0.986}$\spm{0.161}$ & {0.133}$\spm{0.029}$\\
DSMIL     
& {0.699}$\spm{0.032}$ & {0.675}$\spm{0.020}$ & {1.283}$\spm{0.477}$ & {0.145}$\spm{0.065}$\\
TRANS-MIL
& {0.715}$\spm{0.019}$ & {0.686}$\spm{0.036}$ & {1.405}$\spm{0.335}$ & {0.183}$\spm{0.041}$\\
ILRA      
& {0.670}$\spm{0.091}$ & {0.664}$\spm{0.021}$ & {1.186}$\spm{0.296}$ & {0.142}$\spm{0.052}$\\
Slot-MIL  
& {0.684}$\spm{0.071}$ & \textbf{0.703}$\spm{0.023}$ & {1.370}$\spm{0.112}$ & {0.192}$\spm{0.035}$\\
\midrule
\midrule
Slot-MIL + Sub
& {0.722}$\spm{0.043}$ & {0.702}$\spm{0.017}$ & {1.116}$\spm{0.065}$ & {0.144}$\spm{0.016}$\\
Slot-MIL + SubMix
& \textbf{0.733}$\spm{0.047}$ & \textbf{0.703}$\spm{0.017}$ & {0.901}$\spm{0.114}$ & {0.099}$\spm{0.024}$\\
\bottomrule
\end{tabular}
\caption{Multi-class classification results on CAMELYON-17.}
\end{table}

\section{Additional Experiments}
\subsection{Subsampling on Baselines}
\label{app:sec:subonbase}
Subsampling is model-agnostic augmentation method effective to improve generalization performance. We exclude applying subsampling to TRANSMIL, as it relies on the ordering of full patches for the utilization of an additional positional encoding CNN module.

\begin{table}[hbt!]
\centering
\begin{tabular}{l|ccc|ccc}
\toprule
Dataset & \multicolumn{3}{c}{CAMELYON-16} & \multicolumn{3}{c}{TCGA-NSCLC} \\
Method & ACC ($\uparrow$) & AUC ($\uparrow$) & NLL ($\downarrow$) & ACC ($\uparrow$) & AUC ($\uparrow$) & NLL ($\downarrow$)  \\ 
\midrule

ABMIL            & {0.821}$\spm{0.015}$ & {0.808}$\spm{0.034}$ & {1.185}$\spm{0.395}$
                 & {0.832}$\spm{0.039}$ & {0.884}$\spm{0.044}$ & {0.708}$\spm{0.274}$ \\
+ Sub $p=0.1$   
                 & {0.829}$\spm{0.019}$ & {0.812}$\spm{0.046}$ & {0.780}$\spm{0.093}$
                 & {0.852}$\spm{0.021}$ & {0.920}$\spm{0.021}$ & {0.513}$\spm{0.127}$ \\
+ {Sub $p=0.2$}   
                 & {0.843}$\spm{0.010}$ & {0.837}$\spm{0.038}$ & {0.837}$\spm{0.154}$
                 & {0.844}$\spm{0.024}$ & {0.914}$\spm{0.023}$ & {0.572}$\spm{0.123}$ \\
+ {Sub $p=0.4$}   
                 & {0.851}$\spm{0.009}$ & {0.844}$\spm{0.037}$ & {1.015}$\spm{0.207}$
                 & {0.835}$\spm{0.027}$ & {0.899}$\spm{0.027}$ & {0.640}$\spm{0.103}$ \\

\midrule
DSMIL            & {0.839}$\spm{0.012}$ & {0.833}$\spm{0.063}$ & {1.620}$\spm{1.145}$
                 & {0.831}$\spm{0.022}$ & {0.897}$\spm{0.021}$ & {0.737}$\spm{0.098}$ \\
+ {Sub $p=0.1$}   
                 & {0.853}$\spm{0.021}$ & {0.851}$\spm{0.046}$ & {0.780}$\spm{0.287}$
                 & {0.853}$\spm{0.026}$ & {0.922}$\spm{0.022}$ & {0.598}$\spm{0.197}$ \\
+ {Sub $p=0.2$}   
                 & {0.866}$\spm{0.022}$ & {0.870}$\spm{0.046}$ & {0.730}$\spm{0.278}$
                 & {0.850}$\spm{0.032}$ & {0.919}$\spm{0.024}$ & {0.623}$\spm{0.173}$ \\
+ {Sub $p=0.4$}   
                 & {0.848}$\spm{0.031}$ & {0.851}$\spm{0.052}$ & {0.750}$\spm{0.211}$
                 & {0.845}$\spm{0.025}$ & {0.913}$\spm{0.023}$ & {0.617}$\spm{0.093}$ \\
\midrule
ILRA             & {0.830}$\spm{0.010}$ & {0.842}$\spm{0.051}$ & {1.157}$\spm{0.396}$
                 & {0.837}$\spm{0.020}$ & {0.901}$\spm{0.028}$ & {0.824}$\spm{0.117}$ \\      
+ {Sub $p=0.1$}   
                 & {0.842}$\spm{0.011}$ & {0.855}$\spm{0.018}$ & {0.908}$\spm{0.208}$
                 & {0.849}$\spm{0.009}$ & {0.913}$\spm{0.010}$ & {0.852}$\spm{0.187}$ \\
+ {Sub $p=0.2$}   
                 & {0.855}$\spm{0.011}$ & {0.869}$\spm{0.012}$ & {0.825}$\spm{0.268}$
                 & {0.846}$\spm{0.021}$ & {0.916}$\spm{0.022}$ & {0.820}$\spm{0.233}$ \\
+ {Sub $p=0.4$}   
                 & {0.865}$\spm{0.005}$ & {0.895}$\spm{0.012}$ & {0.876}$\spm{0.220}$
                 & {0.831}$\spm{0.021}$ & {0.899}$\spm{0.023}$ & {0.880}$\spm{0.259}$ \\
\midrule
\midrule
Slot-MIL
                & {0.834}$\spm{0.047}$ & {0.893}$\spm{0.023}$ & {1.242}$\spm{0.979}$
                & {0.852}$\spm{0.025}$ & {0.914}$\spm{0.016}$ & {1.001}$\spm{0.202}$ \\
+ {Sub $p=0.1$}   
                 & {0.869}$\spm{0.016}$ & {0.905}$\spm{0.022}$ & {0.508}$\spm{0.043}$
                 & {0.870}$\spm{0.017}$ & {0.929}$\spm{0.019}$ & {0.789}$\spm{0.326}$ \\
+ {Sub $p=0.2$}   
                 & {0.873}$\spm{0.021}$ & {0.911}$\spm{0.021}$ & {0.600}$\spm{0.093}$
                 & {0.873}$\spm{0.020}$ & {0.931}$\spm{0.019}$ & {0.751}$\spm{0.333}$ \\
+ {Sub $p=0.4$}   
                 & {0.881}$\spm{0.024}$ & {0.919}$\spm{0.022}$ & {0.731}$\spm{0.286}$
                 & {0.873}$\spm{0.020}$ & {0.931}$\spm{0.023}$ & {0.866}$\spm{0.313}$ \\
\bottomrule
\end{tabular}
\caption{With varying subsampling rate $p$, Slot-MIL shows robust performance gain compared to other baselines. Experiments are done with ResNet-based features.}
\label{app:tab:aug_compare}
\end{table}

\subsection{Comparing SubMix and RankMix}
\label{appendix:sec:woteacher}
RankMix, built on our Slot-MIL along with patch-level mixup, displays inferior performance compared to SubMix, underscoring the importance of mixing at the attention-based clustered feature level. Without self-training, which is more fair comparison in terms of time complexity, SubMix overwhelms RankMix. All the experiments in \cref{app:tab:subvsrank} are done above the \emph{Slot-MIL}.

\begin{table}[hbt!]
\centering
\begin{tabular}{l|ccc|ccc}
\toprule
Dataset & \multicolumn{3}{c}{CAMELYON-16} & \multicolumn{3}{c}{TCGA-NSCLC} \\
Method & ACC ($\uparrow$) & AUC ($\uparrow$) & NLL ($\downarrow$) & ACC ($\uparrow$) & AUC ($\uparrow$) & NLL ($\downarrow$) \\
\midrule
{RankMix}
                 & {0.870}$\spm{0.023}$ & {0.914}$\spm{0.025}$ & {0.525}$\spm{0.087}$
                 & {\textbf{0.873}}$\spm{0.018}$ & {\textbf{0.932}}$\spm{0.021}$ & {0.532}$\spm{0.173}$ \\
{RankMix} w/o self-training
                 & {0.856}$\spm{0.015}$ & {0.883}$\spm{0.035}$ & {0.550}$\spm{0.103}$
                 & {0.860}$\spm{0.030}$ & {0.926}$\spm{0.030}$ & {0.542}$\spm{0.194}$ \\
{SubMix}
                 & \cellcolor{LightCyan}{\textbf{0.890}}$\spm{0.020}$ 
                 & \cellcolor{LightCyan}{\textbf{0.921}}$\spm{0.020}$ 
                 & \cellcolor{LightCyan}{\textbf{0.448}}$\spm{0.103}$
                 & \cellcolor{LightCyan}{\textbf{0.873}}$\spm{0.023}$ 
                 & \cellcolor{LightCyan}{0.929}$\spm{0.019}$ 
                 & \cellcolor{LightCyan}{\textbf{0.511}}$\spm{0.141}$ \\

\bottomrule
\end{tabular}%
\caption{Without self-training, SubMix works way better than RankMix.}
\label{app:tab:subvsrank}
\end{table}

\subsection{Full results with ResNet-based features}
Due to space constraint, we can't include \gls{acc} metric for CAMELYON-16, and TCGA-NSCLC in \cref{main:tab:main}. In \cref{app:tab:resnetacc}, we provide \gls{acc}.

\begin{table}[hbt!]
\centering
\begin{tabular}{l|ccc|ccc}
\toprule
Dataset & \multicolumn{3}{c}{CAMELYON-16} & \multicolumn{3}{c}{TCGA-NSCLC} \\
Method & ACC ($\uparrow$) & AUC ($\uparrow$) & NLL ($\downarrow$) & ACC ($\uparrow$) & AUC ($\uparrow$) & NLL ($\downarrow$)  \\
\midrule
Meanpool         & {0.664}$\spm{0.017}$ & {0.522}$\spm{0.036}$ & {0.971}$\spm{0.064}$
                 & {0.751}$\spm{0.012}$ & {0.798}$\spm{0.025}$ & {0.571}$\spm{0.047}$ \\ 
Maxpool          & {0.809}$\spm{0.005}$ & {0.783}$\spm{0.022}$ & {0.942}$\spm{0.258}$
                 & {0.739}$\spm{0.018}$ & {0.802}$\spm{0.011}$ & {0.572}$\spm{0.023}$ \\
ABMIL            & {0.821}$\spm{0.015}$ & {0.808}$\spm{0.034}$ & {1.185}$\spm{0.395}$
                 & {0.832}$\spm{0.039}$ & {0.884}$\spm{0.044}$ & {0.708}$\spm{0.274}$ \\
DSMIL            & {0.839}$\spm{0.012}$ & {0.833}$\spm{0.063}$ & {1.620}$\spm{1.145}$
                 & {0.831}$\spm{0.022}$ & {0.897}$\spm{0.021}$ & {0.737}$\spm{0.098}$ \\
TRANSMIL         & {0.813}$\spm{0.029}$ & {0.834}$\spm{0.036}$ & {1.654}$\spm{0.326}$
                 & {0.835}$\spm{0.024}$ & {0.893}$\spm{0.021}$ & {1.791}$\spm{0.559}$ \\
ILRA             & {0.830}$\spm{0.010}$ & {0.842}$\spm{0.051}$ & {1.157}$\spm{0.396}$
                 & {0.837}$\spm{0.020}$ & {0.901}$\spm{0.028}$ & {0.824}$\spm{0.117}$ \\

Slot-MIL        & \cellcolor{LightCyan}{0.834}$\spm{0.047}$ 
                & \cellcolor{LightCyan}{0.893}$\spm{0.023}$ 
                & \cellcolor{LightCyan}{1.242}$\spm{0.979}$
                & \cellcolor{LightCyan}{0.852}$\spm{0.025}$ 
                & \cellcolor{LightCyan}{0.914}$\spm{0.016}$ 
                & \cellcolor{LightCyan}{1.001}$\spm{0.202}$ \\
\midrule
\midrule
DTFD-MIL
                 & {0.847}$\spm{0.016}$ & {0.844}$\spm{0.052}$ & {1.014}$\spm{0.255}$
                 & {0.834}$\spm{0.034}$ & {0.893}$\spm{0.030}$ & {0.980}$\spm{0.178}$ \\
RankMix
                 & {0.870}$\spm{0.023}$ & {0.914}$\spm{0.025}$ & {0.525}$\spm{0.087}$
                 & \textbf{0.873}$\spm{0.018}$ & \textbf{0.932}$\spm{0.021}$ & {0.532}$\spm{0.173}$ \\
SubMix
                 & \cellcolor{LightCyan}\textbf{0.890}$\spm{0.020}$ 
                 & \cellcolor{LightCyan}\textbf{0.921}$\spm{0.020}$ 
                 & \cellcolor{LightCyan}\textbf{0.448}$\spm{0.103}$
                 & \cellcolor{LightCyan}\textbf{0.873}$\spm{0.023}$ 
                 & \cellcolor{LightCyan}{0.929}$\spm{0.019}$ & \cellcolor{LightCyan}\textbf{0.511}$\spm{0.141}$ \\

\bottomrule
\end{tabular}%
\caption{Full results using ResNet-based features}
\label{app:tab:resnetacc}
\end{table}

\subsection{Full results with SimCLR-based features}
\label{app:sec:simclr}
Results with SimCLR features suggest that a \gls{ssl} encoder can enhance performance across \gls{mil} methods. While the gain may seem modest, but given that most models already show relatively high score, any improvement may be regarded as significant. One might wonder why not using SimCLR-based feature for all experiments. However, as extracting contrastive based feature might takes 4 days using 16 Nvidia V100 Gpus \citep{xiang2022exploring} and 2 months to be well-optimized \citep{li2021dual}, it may not be applicable to all real-world scenarios. Given the challenging tasks like C17 multi-class classification, the combination of \gls{ssl}-based features and MIL methods will become crucial. In such demanding scenarios, out method will play significant roles orthogonal to feature extraction method.

\begin{table}[hbt!]
\centering
\begin{tabular}{l|ccc|ccc}
\toprule
Dataset & \multicolumn{3}{c}{CAMELYON-16} & \multicolumn{3}{c}{TCGA-NSCLC} \\
Method & ACC ($\uparrow$) & AUC ($\uparrow$) & NLL ($\downarrow$) & ACC ($\uparrow$) & AUC ($\uparrow$) & NLL ($\downarrow$) \\ 
\midrule
Meanpool         & {0.693}$\spm{0.000}$ & {0.604}$\spm{0.003}$ & {0.674}$\spm{0.023}$
                 & {0.927}$\spm{0.014}$ & {0.972}$\spm{0.010}$ & \textbf{0.232}$\spm{0.055}$ \\
Maxpool          & {0.920}$\spm{0.002}$ & {0.967}$\spm{0.002}$ & {0.353}$\spm{0.147}$          
                 & {0.920}$\spm{0.023}$ & {0.961}$\spm{0.013}$ & {0.608}$\spm{0.053}$ \\
ABMIL            & {0.921}$\spm{0.009}$ & {0.972}$\spm{0.002}$ & {0.234}$\spm{0.033}$
                 & {0.933}$\spm{0.018}$ & \textbf{0.981}$\spm{0.010}$ & {0.263}$\spm{0.101}$ \\ 
DSMIL            & {0.916}$\spm{0.012}$ & {0.968}$\spm{0.008}$ & {0.456}$\spm{0.182}$
                 & {0.936}$\spm{0.017}$ & \textbf{0.981}$\spm{0.010}$ & {0.324}$\spm{0.133}$ \\
TRANSMIL         & {0.889}$\spm{0.026}$ & {0.939}$\spm{0.010}$ & {0.988}$\spm{0.188}$
                 & {0.924}$\spm{0.020}$ & {0.974}$\spm{0.009}$ & {0.381}$\spm{0.127}$ \\
ILRA             & \textbf{0.923}$\spm{0.009}$ & {0.973}$\spm{0.007}$ & {0.333}$\spm{0.043}$
                 & {0.933}$\spm{0.018}$ & \textbf{0.981}$\spm{0.011}$ & {0.277}$\spm{0.103}$ \\
Slot-MIL          
                 & \cellcolor{LightCyan}{0.922}$\spm{0.008}$ 
                 & \cellcolor{LightCyan}{0.972}$\spm{0.007}$ 
                 & \cellcolor{LightCyan}{0.294}$\spm{0.065}$
                 & \cellcolor{LightCyan}\textbf{0.937}$\spm{0.018}$ 
                 & \cellcolor{LightCyan}\textbf{0.981}$\spm{0.011}$ 
                 & \cellcolor{LightCyan}{0.276}$\spm{0.131}$ \\
\midrule
\midrule
SubMix
                 & \cellcolor{LightCyan}\textbf{0.923}$\spm{0.009}$ 
                 & \cellcolor{LightCyan}\textbf{0.975}$\spm{0.008}$ 
                 & \cellcolor{LightCyan}\textbf{0.229}$\spm{0.071}$
                 & \cellcolor{LightCyan}{0.935}$\spm{0.018}$ 
                 & \cellcolor{LightCyan}\textbf{0.981}$\spm{0.012}$ 
                 & \cellcolor{LightCyan}{0.248}$\spm{0.105}$ \\

\bottomrule
\end{tabular}
\caption{Full results using SimCLR-based features}
\label{appendix:tab:tcga_sim}
\end{table}


\subsection{MC inference}
\label{app:sec:mcinference}
\gls{mc} inference performance is measured by averaging $K$ predictions for randomly subsampled patches of a \gls{wsi}. We observed that a value of $K$ less than $100$ leads to a decline in performance. Further details can be found in \cref{appendix:tab:mcinference}. With $K$ over $100$, \gls{mc} inference performance quite matches with full patch inference performance. Although \gls{mc} inference gets better-calibrated prediction, it is not recommended considering the complexity.

\begin{table}[hbt!]
\centering
\resizebox{\linewidth}{!}{
\begin{tabular}{l|ccc|ccc|ccc}
\toprule
Model/Method & & Full Patch & & & MC ($K$=10) & & & MC ($K$=100) & \\
& ACC ($\uparrow$) & AUC ($\uparrow$) & NLL ($\downarrow$) & ACC ($\uparrow$) & AUC ($\uparrow$) & NLL ($\downarrow$) & ACC ($\uparrow$) & AUC ($\uparrow$) & NLL ($\downarrow$) \\ 
\midrule

Sub ($p=0.2$)   & {0.873}$\spm{0.021}$ & {0.911}$\spm{0.021}$ & {0.600}$\spm{0.093}$    
                & {0.843}$\spm{0.014}$ & {0.888}$\spm{0.019}$ & {0.538}$\spm{0.079}$
                & {0.852}$\spm{0.022}$ & {0.895}$\spm{0.020}$ & {0.526}$\spm{0.096}$ \\
Sub ($p=0.4$)   & {0.881}$\spm{0.024}$ & {0.919}$\spm{0.022}$ & {0.731}$\spm{0.286}$
                & {0.860}$\spm{0.025}$ & {0.911}$\spm{0.016}$ & {0.536}$\spm{0.124}$
                & {0.864}$\spm{0.025}$ & {0.913}$\spm{0.020}$ & {0.486}$\spm{0.128}$ \\
SubMix ($p=0.2$)
                & {0.869}$\spm{0.023}$ & {0.905}$\spm{0.026}$ & {0.509}$\spm{0.092}$
                & {0.839}$\spm{0.011}$ & {0.878}$\spm{0.023}$ & {0.494}$\spm{0.050}$
                & {0.848}$\spm{0.017}$ & {0.889}$\spm{0.018}$ & {0.489}$\spm{0.063}$ \\
SubMix ($p=0.4$)
                & {0.890}$\spm{0.020}$ & {0.921}$\spm{0.020}$ & {0.448}$\spm{0.103}$
                & {0.856}$\spm{0.015}$ & {0.903}$\spm{0.015}$ & {0.471}$\spm{0.063}$
                & {0.859}$\spm{0.014}$ & {0.903}$\spm{0.011}$ & {0.473}$\spm{0.077}$ \\
\bottomrule
\end{tabular}
}
\caption{Full Patch inference vs. \gls{mc} inference comparison on CAMELYON-16. Comparisons are made under the same subsampling rate, denoted as $p$. $\alpha=0.5$ for SubMix.}
\label{appendix:tab:mcinference}
\end{table}

\subsection{Ablation on Late Mix}
\label{appendix:sec:latemix}
The late mix ablation is done on Slot-MIL with SubMix augmentation. So the baseline means Slot-MIL + SubMix with $L=0$. We omit this on table for brevity. $\alpha=0.5$ for all experiments. $p=0.2$ for CAMELYON-16, and $p=0.4$ for CAMELYON-17. 

\begin{table}[hbt!]
\centering
\resizebox{0.9\linewidth}{!}{%
\begin{tabular}{l|ccc|ccc}
\toprule
Dataset & \multicolumn{3}{c}{CAMELYON-16} & \multicolumn{3}{c}{CAMELYON-17} \\
Method & ACC ($\uparrow$) & AUC ($\uparrow$) & NLL ($\downarrow$) & ACC ($\uparrow$) & AUC ($\uparrow$) & NLL ($\downarrow$)  \\ 
\midrule
Slot-MIL (Baseline)         
                 & {0.867}$\spm{0.011}$ & {0.908}$\spm{0.016}$ & {0.572}$\spm{0.074}$
                 & {0.785}$\spm{0.060}$ & {0.804}$\spm{0.036}$ & {0.829}$\spm{0.202}$ \\
Slot-MIL + $L=0.1$
                 & {0.872}$\spm{0.022}$ & {0.907}$\spm{0.017}$ & {0.509}$\spm{0.072}$
                 & {0.793}$\spm{0.004}$ & {0.831}$\spm{0.020}$ & {0.641}$\spm{0.014}$ \\
Slot-MIL + $L=0.2$
                 & {0.890}$\spm{0.020}$ & {0.921}$\spm{0.020}$ & {0.448}$\spm{0.103}$
                 & {0.805}$\spm{0.016}$ & {0.835}$\spm{0.015}$ & {0.633}$\spm{0.035}$ \\
Slot-MIL + $L=0.3$
                 & {0.881}$\spm{0.019}$ & {0.917}$\spm{0.019}$ & {0.497}$\spm{0.078}$
                 & {0.799}$\spm{0.012}$ & {0.833}$\spm{0.010}$ & {0.649}$\spm{0.040}$ \\
\bottomrule
\end{tabular}%
}
\caption{Starting mixup from initial epoch is not recommended especially in CAMELYON-16, and CAMELYON-17.}
\end{table}

\section{Experiment Details}
\subsection{Dataset}
\label{app:sec:dataset}
\textbf{TCGA-NSCLC} consists of 528 LUAD and 514 LUSC \glspl{wsi} excluding low-quality ones, following the same experimental setting in DSMIL \citep{li2021dual}. We divide a \gls{wsi} into 224 $\times$ 224-sized patches on 20$\times$ magnification. Then, we extract features on the ResNet-18 model~\citep{he2016deep} pretrained with ImageNet~\citep{deng2009imagenet} available at {timm}~\citep{rw2019timm}. Therefore, the dimension of the extracted feature of each patch is 512. For \textbf{CAMELYON-16}, we use the same features used in \citet{zhang2022dtfd}. Specifically, we split a \gls{wsi} into 256 $\times$ 256-pixel patches on 20$\times$ magnification, while extracting features using ImageNet-pretrained {ResNet-50}~\citep{he2016deep}. The dimension for each patch is 1024. Before passing features through a model, we reduce its dimension to 512 using a linear layer. We use this setting for all models. \textbf{CAMELYON-17} is extracted from ResNet-18 on 224 $\times$ 224-sized patches using 20$\times$ magnification. For the train set, we use \glspl{wsi} scanned from CWZ, RST, and RUMC center. The remaining ones are used for the test set following the standard protocol.

\subsection{Baselines}
\label{app:sec:baseline}
We follow the structure and hyperparameters of original papers, unless otherwise mentioned. For ABMIL, we use gated attention as it performs better than na\"ive attention. For ILRA, we set rank=64, iteration=4, and changed hidden dimension to 128 as it performs better than 256. For DTFD-MIL, we use 5 pseudo-bags for TCGA-NSCLC, and 8 pseudo-bags for CAMELYON-16 and CAMELYON-17. We report the best performance between the two proposed methods: AFS and MaxMinS. For RankMix, we report the best performance within $\alpha \in \{0.5, 1\}$. 

\subsection{Implementation Details}
\label{app:sec:impledetail}
We train the model with 100 epochs for TCGA-NSCLC and 200 epochs for the other datasets. We use {CosineAnnealingWarmRestart}~\citep{loshchilov2016sgdr} with five restarts. We also use {the Adam optimizer}~\citep{kingma2014adam} and set the weight decay as $10^{-4}$ and $(\beta_1, \beta_2)$ as (0.9, 0.999). The batch size is 1 following the conventions.

\end{document}